\documentclass[letterpaper, 10 pt, conference]{ieeeconf}  %

\IEEEoverridecommandlockouts                              %

\usepackage{times}
\usepackage{epsfig}
\usepackage{graphicx}
\usepackage{amsmath}
\usepackage{amssymb}
\usepackage{xspace}
\usepackage{booktabs}
\usepackage{mathtools}

\usepackage{amsfonts}

\usepackage{epstopdf}
\usepackage{color}

\usepackage{multirow}
\usepackage{pbox}

\usepackage{amsopn}

\usepackage{subcaption}
\usepackage{multirow}
\usepackage{longtable}
\usepackage{array}
\usepackage{makecell}

\usepackage{tikz}

\usepackage{cellspace}
\usepackage{hyperref}

\makeatletter

\DeclareRobustCommand\onedot{\futurelet\@let@token\@onedot}
\def\@onedot{\ifx\@let@token.\else.\null\fi\xspace}

\def\eg{\emph{e.g}\onedot} 
\def\ie{\emph{i.e}\onedot} 
 
\def\etc{\emph{etc}\onedot} \def\vs{\emph{vs}\onedot}
 
\def\etal{\emph{et al}\onedot}

\newcommand{\PAR}[1]{\vskip4pt \noindent {\bf #1~}}
\newcommand{\PARbegin}[1]{\noindent {\bf #1~}}

\newcommand{\categoryname}[1]{\textit{#1}}

\newcommand{\fdgvt} {4D-GVT }  %
\newcommand{\mprcnn} {MP R-CNN } %
\newcommand{\mrcnn} {Mask R-CNN } %
\newcommand{\maskxrcnn}{Mask\textsuperscript{X} R-CNN }
\newcommand{\known}{\textit{known} }
\newcommand{\unknown}{\textit{unknown} }
\newcommand{\car}{\textit{car} }
\newcommand{\pedestrian}{\textit{pedestrian} }

\newcommand{\cbox}[1]{\tikz[baseline=-0.5ex]\draw[#1, line width=3, ](0,0) -- (0.2, 0);}

\definecolor{cmprcnn}{RGB}{178, 34, 34}
\definecolor{cagnostic}{RGB}{44, 160, 44}
\definecolor{cthreed}{RGB}{255, 127, 14}
\definecolor{csharpmask}{RGB}{44, 160, 44} %
\definecolor{cmxrcnn}{RGB}{31, 119, 180}
\definecolor{cmrcnn}{RGB}{188, 189, 34}

\definecolor{ccoselect}{RGB}{214, 39, 40}
\definecolor{cfourdgvt}{RGB}{255, 165, 0}
\definecolor{cmxrcnn2}{RGB}{31, 119, 180}
\definecolor{ccamot}{RGB}{139,69,19}

\definecolor{cmaskconsistency}{RGB}{44, 160, 44}
\definecolor{cmotion}{RGB}{255, 165, 0}

\newcommand{\propfig}[4]{
    \includegraphics[width=#1\linewidth]{results/#2/proposals/#3/#4.pdf}
}

\DeclareMathOperator{\pos}{\mathbf{p}}
\DeclareMathOperator{\vel}{\mathbf{v}}
\DeclareMathOperator{\siz}{\mathbf{s}}
\DeclareMathOperator{\cat}{\mathbf{c}}
\DeclareMathOperator{\m}{\mathbf{m}}
\DeclareMathOperator{\score}{s}

\newcommand{\track}[2]{\ensuremath{\operatorname{\mathcal{T}_{#1}^{#2}}}}
\newcommand{\inlierset}[2]{\ensuremath{\operatorname{\mathcal{I}_{#1}^{#2}}}}

\newcommand{\proposal}[2]{\ensuremath{\operatorname{\mathcal{S}_{#1}^{#2}}}}
\newcommand{\proposalset}[2]{\ensuremath{\operatorname{\mathcal{P}_{#1}^{#2}}}}

\title{\LARGE \bf
4D Generic Video Object Proposals
}

\author{Aljo\u{s}a O\u{s}ep, Paul Voigtlaender, Mark Weber, Jonathon Luiten, and Bastian Leibe%
\thanks{
\scriptsize Aljo\u{s}a O\u{s}ep is with the Technical University of Munich. All other authors are with the %
 RWTH Aachen University. %
E-mail: {\tt\scriptsize aljosa.osep@tum.de, lastname@vision.rwth-aachen.de, mark.weber1@rwth-aachen.de}}%
}

\begin{document}

\maketitle
\thispagestyle{empty}
\pagestyle{empty}

\begin{abstract}
Many high-level video understanding methods require input in the form of object proposals. 
Currently, such proposals are predominantly generated with the help of neural networks that were trained for detecting and segmenting a set of \categoryname{known} object classes, which limits their applicability to cases where all objects of interest are represented in the training set. 
We propose an approach that can reliably extract spatio-temporal object proposals for both known and unknown object categories from stereo video.
Our 4D Generic Video Tubes (4D-GVT) method
combines motion cues, stereo data, and data-driven object instance segmentation in a probabilistic framework to compute a compact set of video-object proposals that precisely localizes object candidates and their contours in 3D space and time.
\end{abstract}

\section{Introduction}

\begin{figure}[h]
	\centering
	\setlength{\fboxsep}{0.3pt}%
	\includegraphics[width=1.0\linewidth]{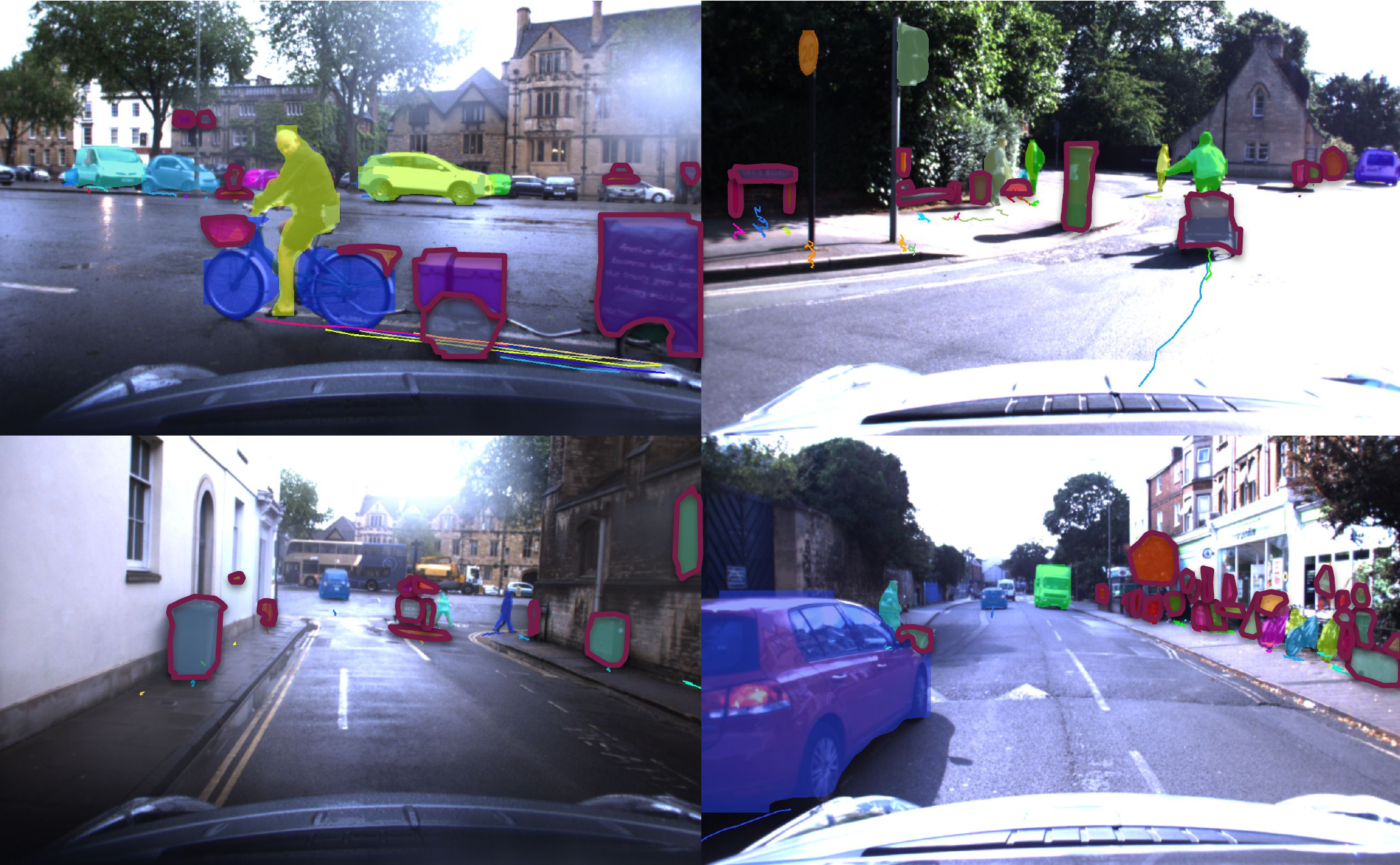}
	\vspace{-15pt}
	\caption{We propose a method for generating 4D video-object tube proposals of arbitrary objects from real-world videos and demonstrate generalization to objects of unknown categories (visualized with a red border).}
	\label{fig:teaser}
\end{figure}

The main result of this paper is a novel approach for generating high-quality spatio-temporal object tube proposals for both \categoryname{known} and \categoryname{unknown} objects from stereo video (Fig.~\ref{fig:teaser}). 
The resulting tube proposals are localized in 3D space and capture the evolution of an object's visible area over time, \ie, they provide a temporally consistent object segmentation over a video sequence. %
Such tube proposals can be useful as basic primitives for a wide variety of applications, ranging from object tracking~\cite{Yoon16CVPR, Choi15ICCV, Kwak15ICCV} to action/activity recognition~\cite{Hou17ICCV}, object category discovery~\cite{Wang14TPAMI, Rubinstein13CVPR, Lee11CVPR, Lee10CVPR, Russell06CVPR, Sivic08CVPR, Zhu12CVPR, Tuytelaars10IJCV} and zero-shot learning~\cite{Zhu19TCSV, Xian18TPAMI, Rahman18ACCV, Bansal18ECCV}. 
4D stereo-based reconstruction and localization in 3D space and time open further possibilities for using \fdgvt as a building block for learning trajectory prediction~\cite{Lee17CVPR, Gupta18CVPR, Alahi16CVPR} and 3D shape completion~\cite{Yuan183DV, Stutz18CVPR}.
These applications are of great importance for automotive applications, in which the capability to perceive and react to unseen dynamic objects is a vital safety concern (see Fig.~\ref{fig:teaser}).

Up to now, high-quality video-object proposals could only be generated either by 1) applying a pre-trained object detector for known classes and performing instance segmentation in every frame~\cite{He17ICCV, Seguin16CVPR}; or by 2) starting from a manual object mask initialization and applying a video object segmentation approach~\cite{Voigtlaender17BMVC, Ting17NIPS}. 
The significant contribution of our approach is that it can provide a compact set of high-quality object tubes for multiple objects with automatic initialization and scales  well to unseen object classes.

The key idea behind our approach is to make use of parallax as a cue to identify temporally consistent object tubes under egomotion of the recording vehicle. 
We leverage recent developments in object instance segmentation~\cite{He17ICCV} and propose a simple, yet effective probabilistic approach for video proposal generation that combines image instance segmentation, stereo, and sparse scene flow cues.
Our proposed \fdgvt method extends Mask R-CNN~\cite{He17ICCV} to extract frame-level object proposals and their segmentation masks for arbitrary objects. 
It then localizes these image regions in 3D space and predicts their 3D motion.
Taking parallax as a consistency filter, our approach narrows down the potentially vast set of tube continuations to those that are consistent with the object's perceived relative motion. 
As a result, \fdgvt can quickly trim down a large initial set of frame-level region proposals and turn them into a temporally consistent set of object tubes with accurately tracked object positions.

Our experiments show that when applying our \fdgvt proposal generator for car and pedestrian tracking on the KITTI dataset~\cite{Geiger12CVPR}, it reaches close to state-of-the-art performance even when compared to dedicated tracking-by-detection methods.
Additionally we compare the obtained results on the image level to those of the Mask\textsuperscript{X} R-CNN, a large-scale instance segmentation  approach by~\cite{Hu18CVPR} that is trained jointly on the COCO~\cite{Lin14ECCV} and Visual Genome~\cite{Krishna16Arxiv} datasets on $3000$ classes. 
Our results show that our approach (4D-GVT) matches accuracy/recall for \categoryname{known} and
\categoryname{unknown} objects, despite only using knowledge about the 80 COCO classes.
All code and data are available at \url{https://github.com/aljosaosep/4DGVT}. %

\section{Related Work}

\PARbegin{Video-Object Mining.} Video-Object mining (VOM) refers to a task of pattern discovery in video collections. 
It has been used for improving detectors by mining hard-negatives for specific object categories from web-videos~\cite{Tang12NIPS, Jin18ECCV}, for learning new detectors of mostly single, dominant objects in videos~\cite{Prest12CVPR} and for tracking-based semi-supervised learning, in which sparse annotations extended by tracks were leveraged to extend the amount of training data~\cite{Misra15CVPR, Misra16ECCV}. 

The above-mentioned methods have in common that they all need to localize video tubes (or object tracks) from video.
Such video tubes or object tracks can be extracted with the help of a pre-trained object detector~\cite{Leibe08TPAMI, Zhang08CVPR}. 
The drawback of such methods is that they can only localize objects, for which a sufficient amount of training data is available. %
In realistic driving scenarios, however, it is not feasible to obtain training data for every possible object of interest. 
For that reason, the most common approach for video-object mining and self-supervised learning in automotive scenarios is based on unsupervised segmentation of 3D sensory data (\eg, LiDAR) using information such as spatial proximity and motion cues~\cite{Dewan15ICRA, Held16RSS, Held14RSS, Teichman12IJRR, Teichman11ICRA}.

There are a few methods that proposed similar ideas in the vision community by using image-based object proposals as leads for tracking~\cite{Zhu16ACCV, Kwak15ICCV, Horbert15ICRA}, often leveraging stereo information~\cite{Osep16ICRA, Osep18ICRA, Mitzel12ECCV, Lenz11IV} or motion cues~\cite{Tokmakov18IJCV}. %
The approach most similar to ours is the category-agnostic multi-object tracker by~\cite{Osep18ICRA}.
It uses two networks, for a) proposal generation and b) track classification, in addition to high-quality dense scene flow~\cite{Vogel13ICCV} for data association.
In contrast, we propose 1) a unified network that provides per-frame mask-level object proposals and a classification branch and 2) a probabilistic framework for offline video-tube generation, derived from well-established MHT theory~\cite{Reid79TAC}. 
Different to~\cite{Osep18ICRA}, our approach models long-term interactions within tubes and utilizes the full video sequence for scoring. As a result, our approach generates tubes that are significantly more stable over time, as our results demonstrate.

\PARbegin{Video-Object Segmentation.} Alternatively, object instances can be mined from video using video-object segmentation (VOS), which refers to a task of segmenting objects in videos with weak supervision, typically in the form of pixel masks or bounding boxes:~\cite{Vondrick18ECCV, Voigtlaender17BMVC, Ting17NIPS, Brendel12ICCV}. Such approaches were successfully used in the context of video-based semi-supervised and self-supervised learning~\cite{Misra15CVPR, Misra16ECCV, Prest12CVPR}.
Existing methods for unsupervised video-object segmentation are limited to segmenting a single object of interest (\ie performing object \vs background classification)~\cite{Fragkiadaki15CVPR, Tokmakov17ICCV, Li18CVPR, Li18CVPR}. 
Both, weakly-supervised and unsupervised VOS methods assume that the object of interest appears from the beginning to the end of the video, with possible short occlusions. 
In contrast, we assume real-world automotive and robotics scenarios, where such assumptions do not hold.

\PARbegin{Open-set Segmentation.} Many real-world scenarios require reliable segmentation that is not limited to a few categories for which fully labeled training data is available. Mask\textsuperscript{X} R-CNN~\cite{Hu18CVPR} tackles this problem by using partial supervision of bounding boxes to perform instance segmentation for the Visual Genome dataset that includes labels for over $3000$ categories. 
Pham \etal~\cite{Pham18ECCV} address the problem by fusing modern instance segmentation with traditional unsupervised methods to obtain instance segmentation for an unbounded number of categories. 
In contrast, our model leverages its in-built category-agnostic detection to generalize to objects of \unknown categories on a video-level.

\section{Method}

\begin{figure}[ht]
	\centering%
	\includegraphics[width=0.7\linewidth]{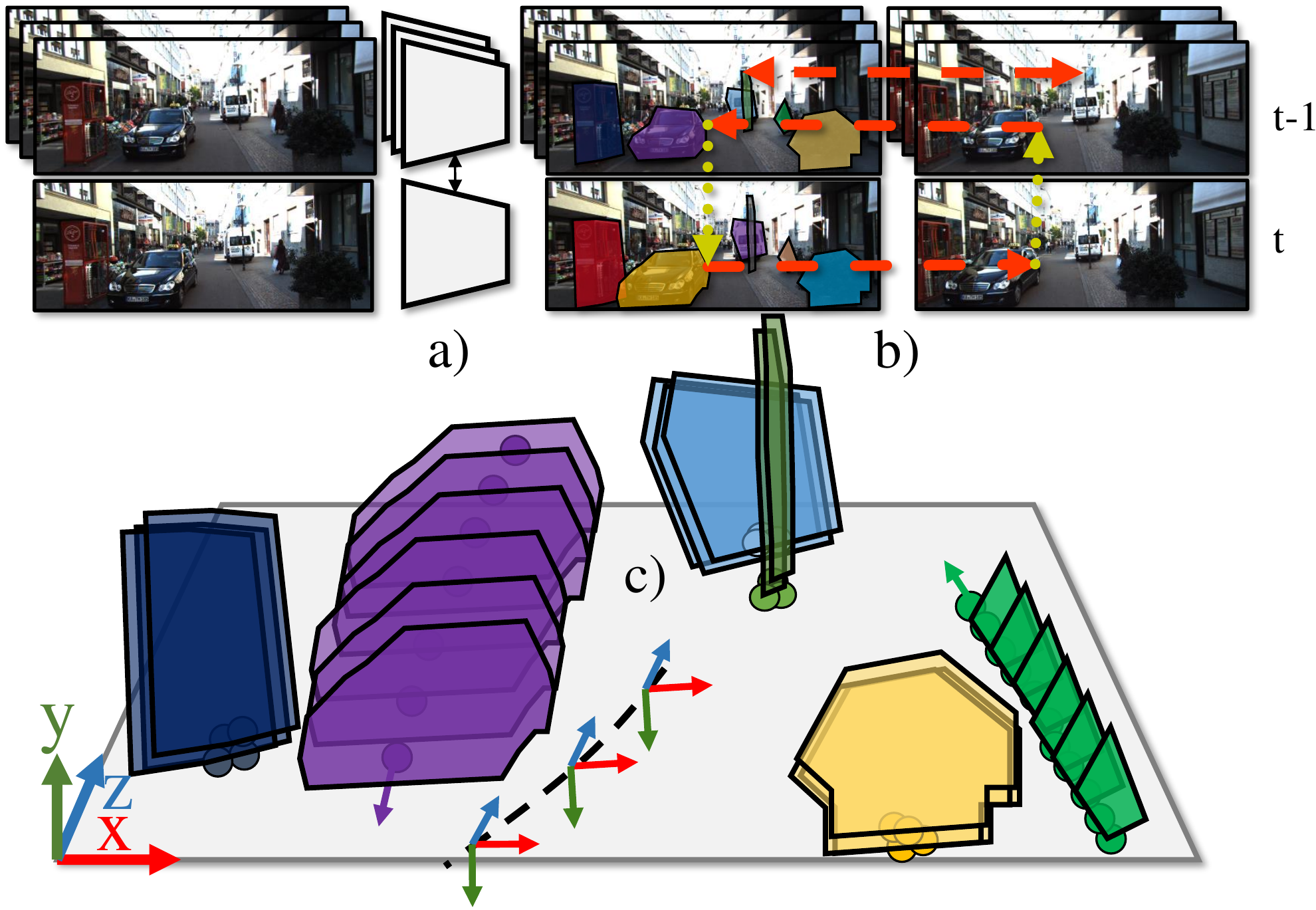}
	\vspace{-7pt}
	\caption{We obtain 4D Video Tubes by a) extracting frame-level object proposals from a video stream using a neural network, b) matching features in order to estimate depth and sparse scene flow to localize proposals in 3D and estimate their velocity and finally, c) associating these 3D segments to obtain a set of video tubes.}
	\label{fig:pipeline}
\end{figure}

From each video-frame of the video sequence, we extract image-level object proposals (Fig.~\ref{fig:pipeline}a), defined by object pixel masks. 
Next, we localize these mask proposals in 3D camera space and predict their motion by computing sparse scene flow, obtained by matching image features (Fig.~\ref{fig:pipeline}b). 
Using the estimated depth and pixel mask, we localize these proposals in 3D space and associate them in space-time in order to obtain a set of 4D Video Proposal Tubes (Fig.~\ref{fig:pipeline}c).
We derive a probabilistic model from well-established MHT theory~\cite{Reid79TAC} that fuses objectness scores provided by the proposal network together with motion and temporal mask consistency cues and that performs an additional inference step that suppresses tubes with significant overlap in space-time.
This way, we combine a learning-based method for generic instance segmentation with a powerful prior knowledge about motion/shape consistency and scene geometry in a sound probabilistic framework.
In the following, we provide a detailed outline of each step of our method.

\subsection{Image-Level Proposals with MP R-CNN}
\label{subsec:imlprop}

First, we need to obtain cues for potential objects on the video-frame level.
For this purpose, we propose a category-agnostic extension of Mask R-CNN, Mask Proposal R-CNN (MP R-CNN), whose classification part only disambiguates objects from the background.
We achieve that by training the network in the category-agnostic setting, \ie
by merging all $80$ COCO classes into one ``object'' class.
This way we obtain a better generalization to object categories that are not present in the training set, as confirmed experimentally in Sec.~\ref{sec:experimental}.
In order to retain a capability to perform higher-granularity classification for objects that appear in the training set, we add a second classification head (identical to the classification head of Mask R-CNN) to this network which disambiguates between the $80$ classes. At test time, the category-agnostic classification head is used to provide confidence scores for proposals, and the second classification head is additionally evaluated to provide a posterior distribution over the $80$ classes for each proposal.

\subsection{4D Video Tubes}
\label{subsec:vt}

To get from image-level proposals to 4D Video Tubes, we combine modern instance segmentation methods, trained in a category-agnostic setting, with well-understood tracking theory~\cite{Leibe08TPAMI, Reid79TAC}.
Formally, we obtain a set of frame-level object proposals $\proposalset{}{0:T}$ in the frame-range $0\leq t \leq T$, where 
$\proposalset{}{t} = \{ \proposal{1}{t}, \dots, \proposal{N_t}{t}\}$ from MP R-CNN.
The $j$-th per-frame proposal is defined as $\proposal{j}{t} = [ \pos_j^t, \vel_j^t, \siz_j^t, \cat_j^t, \m_j^t, \score_j^t ]$. 
The mask $\m \in \{0, 1\}^{w \times h}$ for an image of size $w \times h$, objectness score $\score \in [0, 1]$ and category information $\cat \in \mathbb{R}^{81}$ ($80$ \categoryname{known} COCO classes + \categoryname{unknown}) are obtained by MP R-CNN.

This information is extended by 3D position $\pos \in \mathbb{R}^3$, 3D size $\siz \in \mathbb{R}^3$ and velocity vector $\vel \in \mathbb{R}^3$ from the video sequence as follows. 
For each video sequence, we match local Harris corner based image features 1) of left-right images coming from stereo and 2) forward-backward in time for two consecutive neighboring frames $(t-1)$ and  $t$. 
Features that pass a cyclic consistency check (left, right, forward, backward) are used to estimate sparse scene flow. We triangulate left-right matches in two consecutive frames and compute difference vectors ($(t-1) \to t$) for all successful matches. 
Using the same set of feature matches, we obtain an egomotion estimate $E^{(t-1) \to t}$ by minimizing the re-projection error using non-linear least squares~\cite{Geiger11IV} and we finally keep left-right image matches to obtain a scene depth estimate $D^t$ using the feature-matching method of~\cite{Geiger10ACCV}.
By cropping stereo and scene flow estimates within the estimated mask area, we robustly compute the 3D position, velocity and 3D bounding box for each proposal. 

\PAR{Video Tube Proposals.}~After obtaining the per-frame object proposals $\proposalset{}{0:T}$ for a video sequence that spans from frame $0$ to $T$, we partition these into a 4D tube proposal set, \ie, a set of video tube proposals $\mathcal{A}^{0:T}  = \{ \track{i}{n_i:m_i} \vert i=1,\dots,N \}$, where each tube proposal \track{i}{} spans from frame $n_i$ to $m_i$.

Obtaining such partitions is a very challenging, multi-dimensional assignment problem: object proposals may originate from objects, as well as from segmentation clutter.
Obtaining an exact solution to multi-frame assignment problems in tracking is NP-hard, and we need to resort to approximate solutions. 
One possibility would be to use an approach like MHT~\cite{Reid79TAC}, which maintains a tree of potential associations for each possible object and to perform tree pruning in order to avoid combinatorial explosion. 
However, for a large number of per-frame object proposals, maintaining such a tree would be prohibitively expensive.

Instead, we resort to the forward-backward track enumeration algorithm~\cite{Leibe08TPAMI}, guided by the estimated scene flow. 
We perform a forward-backward data association by sliding a temporal window over the whole video sequence. 
For each video frame $t \in \{0, \ldots, T\}$, we 1) extend the existing tube proposal set using $\proposalset{}{t} $ and 2) attempt to start a new object tube from each unassigned proposal by performing bi-directional data association within a temporal window. 

In order to minimize incorrect associations, which would lead to spurious tubes, we leverage object motion models, the rigidity of the 3D scene, and the known image formation model to guide frame-to-frame proposal data association to only a small set of feasible associations. This is done as follows (similar to~\cite{Osep18ICRA}).
Using an estimate of frame-level object proposals with 3D position $\pos$, velocity $\vel$ and size $\siz$ from each frame-level proposal, we start a new tube and estimate its state recursively using a linear Kalman filter. 
We perform association in the following steps. 
First, we perform a Kalman filter prediction in order to estimate a 3D position and perform 3D position based gating in order to severely narrow down the set of feasible proposal associations. 
Then we predict the segmentation mask of the object in the current frame, and finally we extend the tube with the proposal that maximizes the joint association probability based on mask IoU in the image domain and 3D position, as in~\cite{Osep18ICRA}.

We obtain the mask prediction $\tilde{\mathbf{m}}^{t}_i \in \{0, 1\}^{w \times h}$ of tube $\mathcal{T}_i$ for video-frame $ 1 \leq t \leq T$ based on a 2D mask estimate of the last associated mask $\mathbf{m}^{(t-1)}_i$, the depth $D^t$ and egomotion estimate $E^{(t-1)\to t}$:
\begin{equation}
    \footnotesize
    \begin{aligned}
        & \mathcal{J} = \pi (K E^{(t-1)\to t}\ M_i^{(t-1)\to t} K^{-1} \pi^{-1} (\mathbf{m}_i^{(t-1)} \odot D^{(t-1)})) , \\
        & \tilde{\mathbf{m}}^{t}_i ({\mathbf{p}}') = \left\{\begin{matrix}
        1, {\mathbf{p}}' \in \mathcal{J} \\ 
        0, \text{else}\;\;\;\;,
        \end{matrix}\right.
    \end{aligned}
\end{equation}
where $\pi: \mathbb{R}^3 \to \mathbb{R}^2$ and $\pi^{-1}: \mathbb{R}^2 \to \mathbb{R}^3$ denotes the camera projection/backprojection operator (applied element-wise), $K$ denotes the known camera matrix and $\odot$ performs element-wise multiplication. The matrix $M_i^{(t-1) \to t}$ represents estimated motion of the tube by the Kalman filter.
Thus, starting from the per-frame proposals $\proposalset{}{0:T}$ over the video sequence, obtained from the \mprcnn network and localized in 3D space, we have now linked these proposal segments together into a set of proposal video tubes $\mathcal{A}^{0:T}$.

\PAR{Scoring Tube Proposals.}~Following MHT theory~\cite{Reid79TAC}, we obtain a ranked set of relevant video tube proposals by scoring tubes using log-likelihood ratios between the proposal tube and the null hypothesis tube. 
The null tube encodes tubes generated from random background clutter.
The score $\theta (\cdot)$ of the tube $\track{i}{}$ is a linear combination of the tube motion score, the mask consistency score and the tube objectness score (we are omitting time indices for brevity):
\begin{equation}
    \label{equ:tubescore}
     \footnotesize
    \theta (\track{i}{}) = w_1 \cdot \vartheta_{\text{mot}} (\track{i}{}) + w_2 \cdot \vartheta_{\text{mask}} (\track{i}{}) + w_3 \cdot \vartheta_{\text{obj}} (\track{i}{}).
\end{equation}
Intuitively, the \textbf{tube motion score} encodes the assumption that objects move smoothly. It takes the form of a likelihood ratio test, comparing the probability that the tube corresponds to a valid object tube to the null hypothesis that the tube was generated from random background clutter:
\begin{equation}
    \footnotesize
    \vartheta_{\text{mot}} (\track{i}{n:m}) = \ln \frac{p(\pos_{\inlierset{i}{}}^{n:m} | \inlierset{i}{} \subseteq \track{i}{n:m} )}{p(\pos_{\inlierset{i}{}}^{n:m} | \inlierset{i}{} \subseteq \track{\emptyset}{} )}.
    \label{equ:llmotion}
\end{equation}
The notation $\inlierset{i}{} \subseteq \track{i}{n:m}$ denotes that the 3D position observations originate from the tube $\track{i}{n:m}$ and $\track{\emptyset}{}$ denotes the null tube hypothesis.
The set $\inlierset{i}{n:m} = \{ \proposal{k_i}{t} \;|\; n \leq t \leq m  \}$ denotes the supporting image-level proposal set of a tube, where $k_i$ refers to the $k$-th frame-level proposal that supports $\track{i}{}$ at a certain frame $t$.
The Markovian motion assumption and conditional independence assumption of the observations originating from the null tube hypothesis lead to the following factorization of Eq.~\ref{equ:llmotion}:
\begin{equation}
    \footnotesize
    \begin{aligned}
    \begin{split}
    p(\pos_{\inlierset{i}{}}^{n:m} | \inlierset{i}{} \subseteq \mathcal{T}_i^{n:m} )
    &= \smashoperator{\prod_{t=n}^{m}} p (\mathbf{p}^t | \{ \mathbf{p}_{\inlierset{i}{}}^{t^\prime} | n \leq t^\prime < t \} , \inlierset{i}{} \subseteq \mathcal{T}_i^{n:t}) \\
    &= \smashoperator{\prod_{t=n}^{m}} \mathcal{N} (\pos^t; \tilde{\pos}^{t}, \tilde{\Sigma}^{t}),
    \end{split}
    \end{aligned}
\end{equation}
\begin{equation}
    \footnotesize
    p(\pos_{\inlierset{i}{}}^{n:m} | \inlierset{i}{} \subseteq \mathcal{T}_\emptyset ) = \smashoperator{\prod_{t=n}^{m}} p (\mathbf{p}_{\inlierset{i}{}}^t | \inlierset{i}{} \subseteq \track{i}{}) = \smashoperator{\prod_{t=n}^{m}}\ \frac{1}{A_{\text{GP}}}.
\end{equation}
The likelihood that the observation originates from the tube is assumed to be Gaussian and is evaluated using the Kalman filter prediction. 
It measures how well the ground position of an object proposal corresponds to the Kalman filter prediction $\tilde{\pos}^{t}$ under the estimated variance $\tilde{\Sigma}^{t}$ at time $t$.
This is evaluated against the null hypothesis of a uniform distribution of clutter objects over the sensing area $A_{\text{GP}}$ (roughly $80m \times 50m$ of the ground plane). %

The \textbf{tube mask consistency score} encodes the intuition that the silhouette and position of the object in the image plane do not change significantly on a frame-to-frame level.
Again, a likelihood ratio test is used, comparing the mask IoU against a null hypothesis model of no intersection:
\begin{equation}
    \footnotesize
    \vartheta_{\text{mask}} (\track{i}{n:m}) = \ln \frac{p(\m_{\inlierset{i}{}}^{n:m} | \inlierset{i}{} \subseteq \mathcal{T}_i^{n:m} )}{p(\m_{\inlierset{i}{}}^{n:m} | \inlierset{i}{} \subseteq \track{\emptyset}{} )},
    \label{equ:llmask}
\end{equation}
which using the Markov assumption factorizes as:
\begin{equation}
    \footnotesize
    \begin{aligned}
    \begin{split}
    p(\m_{\inlierset{i}{}}^{n:m} | \inlierset{i}{} \subseteq \mathcal{T}_i^{n:m} )
    &= \prod_{t=n}^{m} p (\m^t | \{ \m^{t^\prime}_{\inlierset{i}{}} | n \leq t^\prime < t \}, \inlierset{i}{} \subseteq \mathcal{T}_i^{n:t}) \\
    &= \prod_{t=n}^{m} \text{IoU} (\tilde{\m}^{t}, \m^t)
    \end{split}
    \end{aligned}
\end{equation}
\begin{equation}
    \footnotesize
    p(\m_{\inlierset{i}{}}^{n:m} | \inlierset{i}{} \subseteq \track{\emptyset}{} ) = \prod_{t=n}^{m} p (\m_{\inlierset{i}{}}^t | \inlierset{i}{} \subseteq \track{\emptyset}{}) = \prod_{t=n}^{m} \frac{1}{\alpha},
\end{equation}
Here, we compute frame-to-frame mask consistency as mask intersection-over-union between the mask prediction $\tilde{\m}^{t}$ and the (mask) observation $\m^t$.
The null hypothesis model is a confidence threshold (base value $\alpha$), obtained by hyperparameter optimization.

Finally, the \textbf{tube objectness score} utilizes the objectness scores estimated by the network and integrates them over the whole tube:
\begin{equation}
	\label{eq:seq_objectness}
    \footnotesize
    \vartheta_{\text{obj}} (\track{i}{n:m}) = \ln \frac{p(\score_{\inlierset{i}{}}^{n:m} | \inlierset{i}{} \subseteq \mathcal{T}_i^{n:m} )}{p(\score_{\inlierset{i}{}}^{n:m} | \inlierset{i}{} \subseteq \track{\emptyset}{} )},
\end{equation}
\begin{equation}
    \footnotesize
    \begin{aligned}
    \begin{split}
    p(\score_{\inlierset{i}{}}^{n:m} | \inlierset{i}{} \subseteq \mathcal{T}_i^{n:m} ) = \prod_{t=n}^{m} p (\score^t_{\inlierset{i}{}} | \inlierset{i}{}  \subseteq \mathcal{T}_i^{n:t})
                                                                                       = \prod_{t=n}^{m} \score^t,
    \end{split}
    \end{aligned}
\end{equation}
\begin{equation}
    \footnotesize
    p(\score_{\inlierset{i}{}}^{n:m} | \inlierset{i}{} \subseteq \track{\emptyset}{} ) = \prod_{t=n}^{m} p (\score_{\inlierset{i}{}}^t | \inlierset{i}{} \subseteq \track{\emptyset}{}) = \prod_{t=n}^{m} \frac{1}{\beta}.
\end{equation}
Here, the likelihood ratio test compares the probability of the tube being a valid object (individual probabilities are estimated by the \mprcnn network) against a null hypothesis model of the tube belonging to segmentation clutter. 
The confidence threshold $\beta$ is a hyperparameter.

Experimentally, we can observe (Sec.~\ref{sec:experimental}) that for \categoryname{known} object categories, sequence-level objectness scores (Eq.~\ref{eq:seq_objectness}) provide a great ranking cue. 
However, for objects which do not belong to one of the categories in the training set (80 COCO classes), scores based on temporal mask and motion consistency lead to a significantly better ranking. %

\PAR{Proposal Tubes Co-Selection.}~These proposal tubes may still overlap with each other and provide competing explanations for the same image pixels. 
In order to obtain a compact video tube set, we perform an additional video tube co-selection step that
suppresses highly-overlapping tubes.
The co-selection is performed by computing a maximum a posterior probability (MAP) estimate using a Conditional Random Field (CRF) model by minimizing the following non-submodular function within batches of $100$ frames:
\begin{equation}
    \footnotesize
    \label{equ:energy}
    \text{F} (\mathbf{b}, \mathcal{A}^{n:m}) = \smashoperator{\sum_{\mathcal{T}_i \in \mathcal{A}^{n:m}}} b_i (\epsilon_1 - \theta (\mathcal{T}_i)) + \epsilon_2 \cdot \smashoperator{\sum_{\mathcal{T}_i, \mathcal{T}_j \in \mathcal{A}^{n:m}}} b_i b_j \varphi (\mathcal{T}_i, \mathcal{T}_j)
\end{equation}
$\mathbf{b} \in \{0, 1\}^{|\mathcal{A}^{n:m}|}$ is a binary vector whose elements $b_i$ indicate that the $i^{th}$ tube is selected. 
The unary term $\theta(\cdot)$ corresponds to the tube scoring function.
The pairwise term $\varphi (\mathcal{T}_i, \mathcal{T}_j) = \sum_{t=n}^{m} \ln (\text{IoU}(\mathbf{m}^t_i, \mathbf{m}^t_j))$ integrates image-level temporal mask IoU within the batch and gives a penalty to overlapping tubes.
Here, $\mathbf{m}_i^t$ and $\mathbf{m}_j^t$ are object masks of tubes $\track{i}{}$ and $\track{j}{}$ at frame $t$, respectively. 
The model parameters $\epsilon_1$ and $\epsilon_2$ are optimized on a held-out validation set of the KITTI dataset~\cite{Geiger12CVPR} using random search.

To minimize the energy function (Eq. \ref{equ:energy}), we use the multi-branch algorithm by~\cite{Schindler06ECCV}.

\section{Experimental Evaluation}
\label{sec:experimental}

\PAR{Datasets.}~We evaluate the proposed method on two automotive datasets, KITTI~\cite{Geiger12CVPR} and Oxford RobotCar~\cite{Maddern17IJRR}. 
The KITTI dataset is a standard object detection and tracking benchmark in automotive scenarios. %
We use the KITTI Multi-Object Tracking benchmark with the \categoryname{car} and \categoryname{pedestrian} categories for a sequence-level evaluation of our video tubes.

The Oxford RobotCar dataset provides a large amount of recorded data (roughly $30$h) and covers a broad set of object classes. 
However, there are no object class labels available. 
For evaluation purposes, we labeled a subset of the video sequences (see Tab.~\ref{tab:stats-oxford-labels-recall}) in order to measure the performance of our method on \known and \unknown object classes.

\begin{table}
    \footnotesize
    \begin{center}
    \begin{tabular}{l c c c c c c}
\toprule
Category & Car & Person & Bike & Bus & Other & All\tabularnewline
\hline 
\#instances & 599 & 354 & 78 & 50 & 413 & 1494\tabularnewline
Portion & 40.1\% & 23.1\% & 5.2\% & 3.3\% & 27.6\% & 100\% \tabularnewline
\bottomrule
\end{tabular}
    \end{center}
    \vspace{-10pt}%
    \caption{We labeled a subset of 150 images of different videos of the Oxford RobotCar dataset for evaluation.} %
    \label{tab:stats-oxford-labels-recall}
\end{table}

\begin{figure}[t]
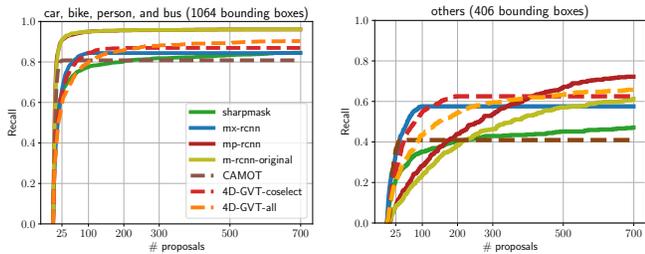

    \centering
    \propfig{0.47}{2019-03-21-16-10-35}{proposals_and_tracklets}{car_bike_person_andbus_modif}
    \propfig{0.47}{2019-03-21-16-10-35}{proposals_and_tracklets}{others}
    \vspace{-18pt}
    \caption{Recall evaluation of image-level (solid) and video tube (dashed) proposals on the Oxford RobotCar dataset. Both plots use the same legend. Remark: (\protect\cbox{cmprcnn}) and (\protect\cbox{cmrcnn}) follow the same curve in the left plot.}
    \label{fig:vom-oxford-baselines}
\end{figure}

\PAR{Evaluation on Oxford.}%
Using our labeled subset of the Oxford RobotCar dataset, we compare \mprcnn and \fdgvt to several baselines. 
All methods were trained on the COCO dataset, except Mask\textsuperscript{X} R-CNN~\cite{Hu18CVPR}, which was trained on both COCO and Visual Genome~\cite{Krishna16Arxiv} with $3000$ object categories.
Fig.~\ref{fig:vom-oxford-baselines} shows recall (at $>\!\!0.5$ IoU) as a function of the number of object proposals for \known object categories (in the COCO category set, \textit{left}) and \unknown object categories (\textit{right}).
For image-level evaluation (solid lines), we compare our \mprcnn (\cbox{cmprcnn}) to SharpMask~\cite{Pinheiro16ECCV} (%
\cbox{csharpmask}), \maskxrcnn (\cbox{cmxrcnn}) and standard Mask R-CNN (\cbox{cmrcnn}) where we merely keep the low-confidence predictions, which is a strong baseline as shown in~\cite{Pham18ECCV}.  
As can be seen in Fig.~\ref{fig:vom-oxford-baselines} (\textit{left}), our \mprcnn model (\cbox{cmprcnn}) and Mask R-CNN (\cbox{cmrcnn}) perform well on \known object categories while SharpMask (\cbox{csharpmask}) and \maskxrcnn (\cbox{cmxrcnn}) achieve lower recall. 
On \unknown object categories \maskxrcnn performs significantly better for top-$100$ proposals per frame and provides a better ranking, which is not surprising, as it was trained on a large dataset, containing $3000$ classes. 
In the limit of $700$ proposals per frame, \mprcnn (\cbox{cmprcnn}) and Mask R-CNN (\cbox{cmrcnn}) achieve higher recall ($0.75$ and $0.61$, respectively) than \maskxrcnn (\cbox{cmxrcnn}) ($0.59$).

We evaluate our \fdgvt tubes (dashed lines) on the Oxford RobotCar dataset by extracting video tubes in a $100$-frame temporal neighborhood of the $150$ labeled images. 
We show the results obtained using the full tube set (\cbox{cfourdgvt} \cbox{cfourdgvt}) and the set obtained after the co-selection step (\cbox{ccoselect} \cbox{ccoselect}), as explained in Sec.~\ref{subsec:vt}.
For \known classes, we achieve comparable recall to \maskxrcnn (\cbox{cmxrcnn}) with both 4D-GVT variants (\cbox{cfourdgvt} \cbox{cfourdgvt}, \cbox{ccoselect} \cbox{ccoselect}) in the top-$100$ proposals regime. 

For \unknown categories, \maskxrcnn (\cbox{cmxrcnn2}) and both 4D-GVT variants provide an excellent ranking for the top-$100$ proposals. 
Remarkably, our 4D-GVT proposals achieve the same recall and, most importantly, ranking capabilities as \maskxrcnn despite being trained only on the $80$ COCO categories. Also supremely important: \fdgvt tracks the objects and assigns temporarily consistent objects ids, which \maskxrcnn cannot do.
This clearly demonstrates that the generalization of Mask R-CNN baseline to \unknown object classes can be significantly improved by leveraging parallax as consistency filter and motion cues.

A comparison to unsupervised video-object segmentation methods such as~\cite{Fragkiadaki15CVPR, Tokmakov17ICCV, Li18CVPR, Li18CVPR} is not possible as these methods extract only a single dominant tube, while the sequences from Oxford RobotCar contain several object instances in every video-frame. 
However, as an additional video-level baseline, we adapt the official implementation of CAMOT~\cite{Osep18ICRA} and replace the two-stage proposal generation and track classification with our MP R-CNN for a fair comparison. 
As can be seen from Fig.~\ref{fig:vom-oxford-baselines} (\textit{left}), CAMOT (\cbox{ccamot} \cbox{ccamot}) can track several \categoryname{unknown} objects, but achieves significantly lower recall ($0.4$ vs. $0.61$ for 4D-GVT-coselect).

Fig.~\ref{fig:qualitative} compares the output of (\textit{a}) our method, (\textit{b}) \maskxrcnn and (\textit{c}) Mask R-CNN qualitatively.
As can be seen, \mrcnn works very well for the $80$ classes (\categoryname{car}, \categoryname{bike}, \categoryname{person}, \etc). \fdgvt and \maskxrcnn both segment a large variety of objects. However, only our method additionally provides temporal continuity.

\begin{figure}[t]
\centering
    \includegraphics[width=0.47\linewidth]{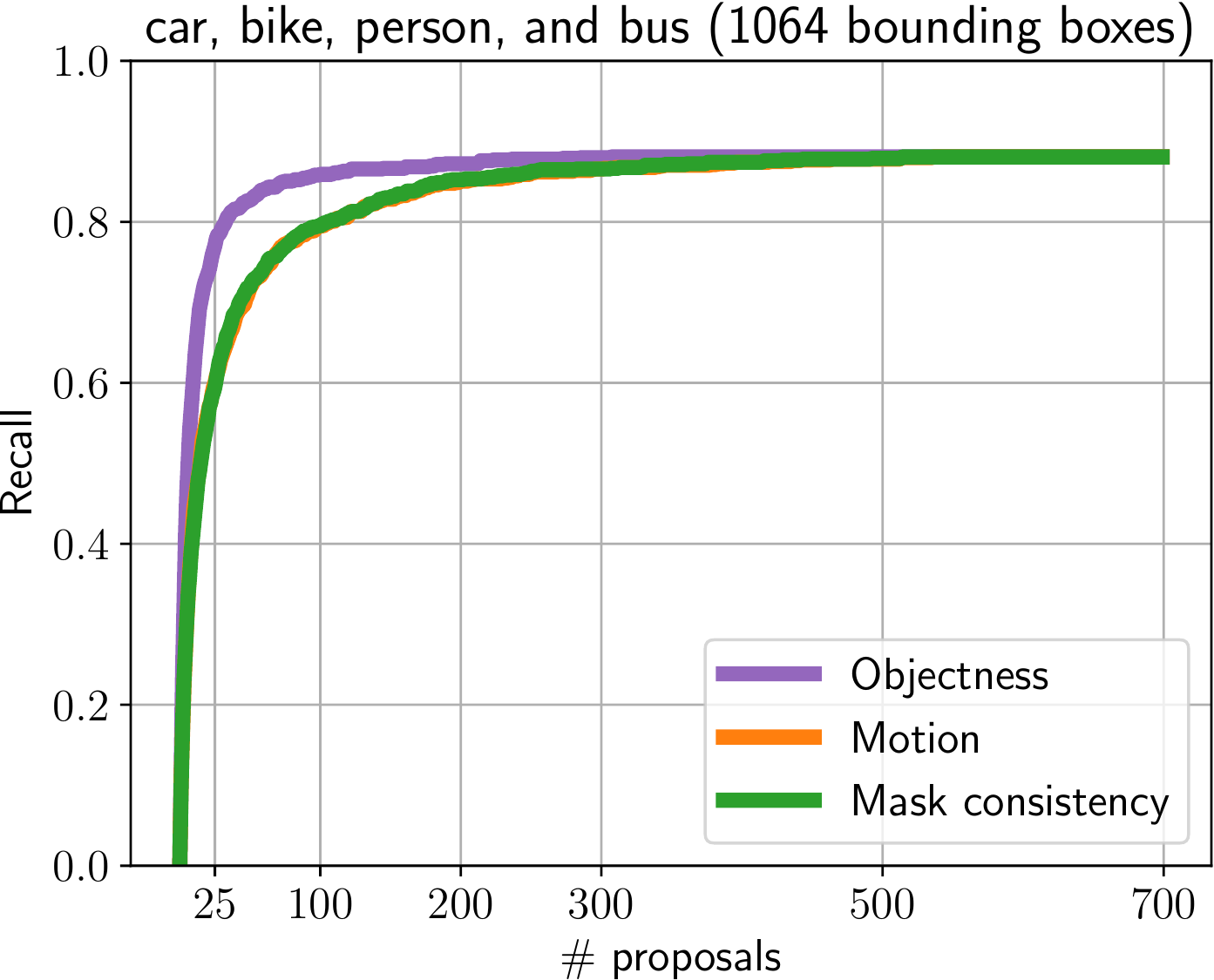}
    \includegraphics[width=0.47\linewidth]{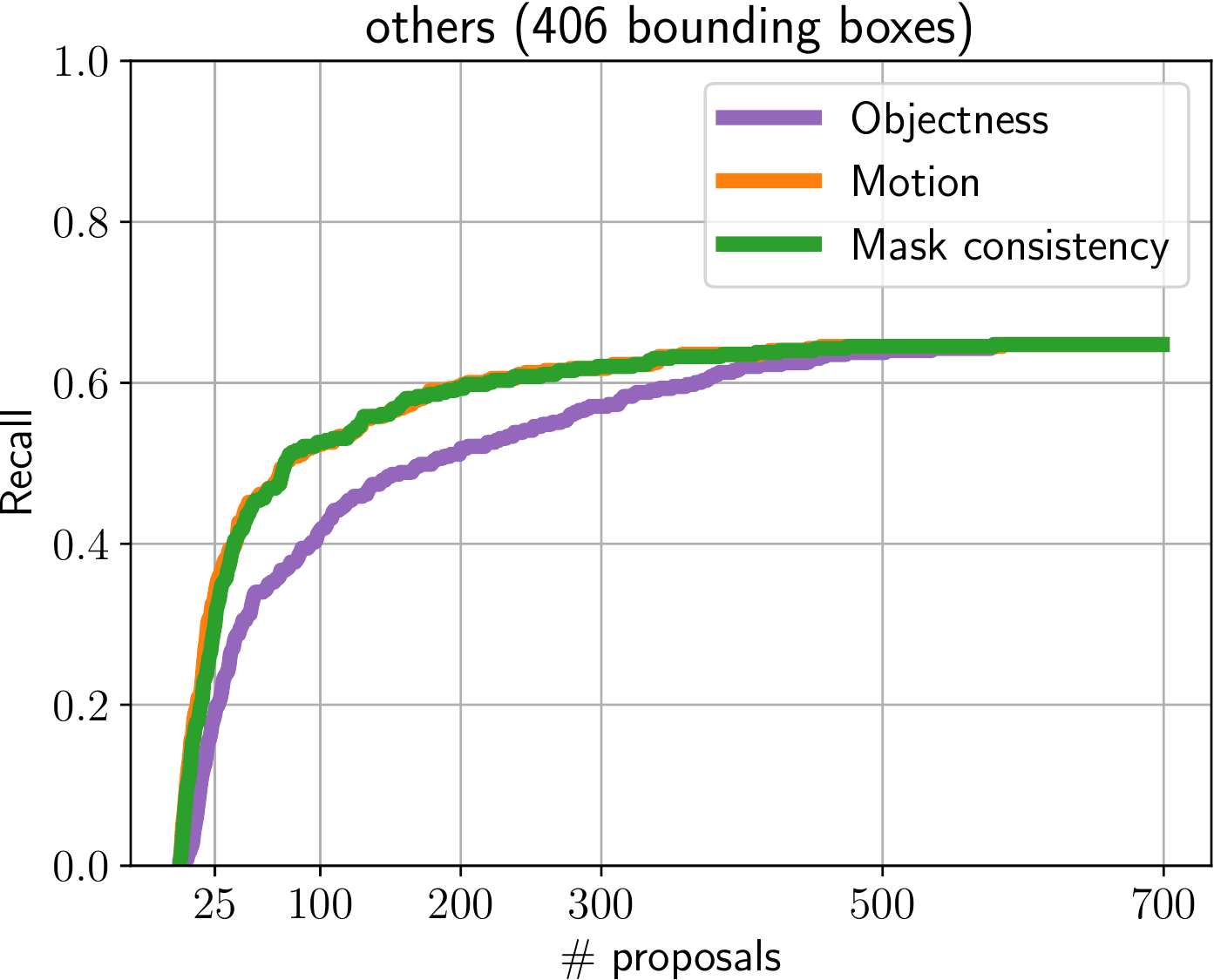}
    \vspace{-6pt}
    \caption{Tube scoring evaluation (Oxford). Remark: (\protect\cbox{cmaskconsistency}) and (\protect\cbox{cmotion}) follow the same curve.}
\label{fig:vom-tracklet-scoring}
\end{figure}

\PAR{Scoring Ablation.}~We experimentally justify the design decisions for tube scoring (Oxford) in Fig.~\ref{fig:vom-tracklet-scoring}. Clearly, for \known object categories, tube ranking based on the proposal confidence scores (\textbf{objectness}) obtained by MP R-CNN is a strong cue (Fig.~\ref{fig:vom-tracklet-scoring}, \textit{left}). 
However, for \unknown categories (Fig.~\ref{fig:vom-tracklet-scoring}, \textit{right}), the scoring based on the \textbf{mask consistency} and the \textbf{motion} scores provide significantly stronger scoring cues. 
This further confirms the efficacy of utilizing prior knowledge about motion and temporal shape consistency to compensate for the lack of training data.

\PAR{Sequence-Level Evaluation.}~We demonstrate sequence-level performance by evaluating our method on our custom split of the training set%
\footnote{We used sequences 1,2,3,4,9,11,12,15,17,19,20 to perform model validation and the rest for the evaluation.} from the KITTI Multi-Object Tracking dataset (KITTI MOT)~\cite{Geiger12CVPR}. 
We follow the official evaluation protocol that evaluates tracking performance using the   CLEAR-MOT~\cite{Bernardin08JIVP} metrics on the \categoryname{car} and \categoryname{pedestrian} categories, as only these categories are observed frequently enough so that a tracking evaluation is statistically meaningful.
In general, we do not know the object categories of the tubes. 
However, we can use the information obtained from the second (category-specific) classification head (as explained in Sec.~\ref{subsec:imlprop}) and obtain the conditional probability for a tube representing a \categoryname{car} or \categoryname{pedestrian}, given that it represents an object.
We compare our method to the tracking-by-detection approach by~\cite{Osep17ICRA} (CIWT), which is among the top-4 performers on the KITTI MOT benchmark (comparing methods using public detections, Regionlets~\cite{Wang13ICCV}) and which thus provides a strong object tracking baseline. 
This method is tuned for tracking \categoryname{car} and \categoryname{pedestrian} categories only and uses dedicated motion models for those categories. 
In addition, we compare to a stereo-based category-agnostic tracker (CAMOT)~\cite{Osep18ICRA} that is capable of tracking a large variety of objects. 
To make the comparison fair, we use the same inputs from MP R-CNN, which is trained on COCO, for CAMOT and our method. 
In this case, we map the COCO category \categoryname{person} to \categoryname{pedestrian}. 
As an additional experiment, we fine-tune MP R-CNN on the KITTI dataset on the \car and \pedestrian categories, which we can use to evaluate CIWT. Both CAMOT and \fdgvt output bounding boxes derived from the tracked segments, whereas KITTI MOT assumes amodal bounding boxes.

\begin{table}[t]
\footnotesize
  \begin{center}
    \begin{tabular}{lrrrr}
	   \toprule
       & \multicolumn{2}{c}{\textbf{Pedestrian}} & \multicolumn{2}{c}{\textbf{Car}} \\
       & MOTA & IDS & MOTA & IDS \\ 
		\midrule   
		CIWT (KITTI) & \textbf{0.33} & 42 & \textbf{0.65} & 8 \\
    CAMOT (KITTI) & 0.26 & 72 & 0.60 & 40 \\
		4D-GVT (KITTI) & \textbf{0.33} & \textbf{18} & 0.61 & \textbf{6}  \\
		\midrule
    CAMOT (COCO) & -0.05 & 214 & 0.54 & 174 \\
		4D-GVT (COCO) & \textbf{0.27} & \textbf{24} & \textbf{0.57} & \textbf{4} \\
		\bottomrule
    	\end{tabular}
    \vspace{-5pt}
    \caption{\small Tracking evaluation on the KITTI MOT dataset.} %
    \label{tab:clearmot-ped}
  \end{center}
\end{table}

\begin{figure}[t]
\centering
    \includegraphics[width=0.49\linewidth]{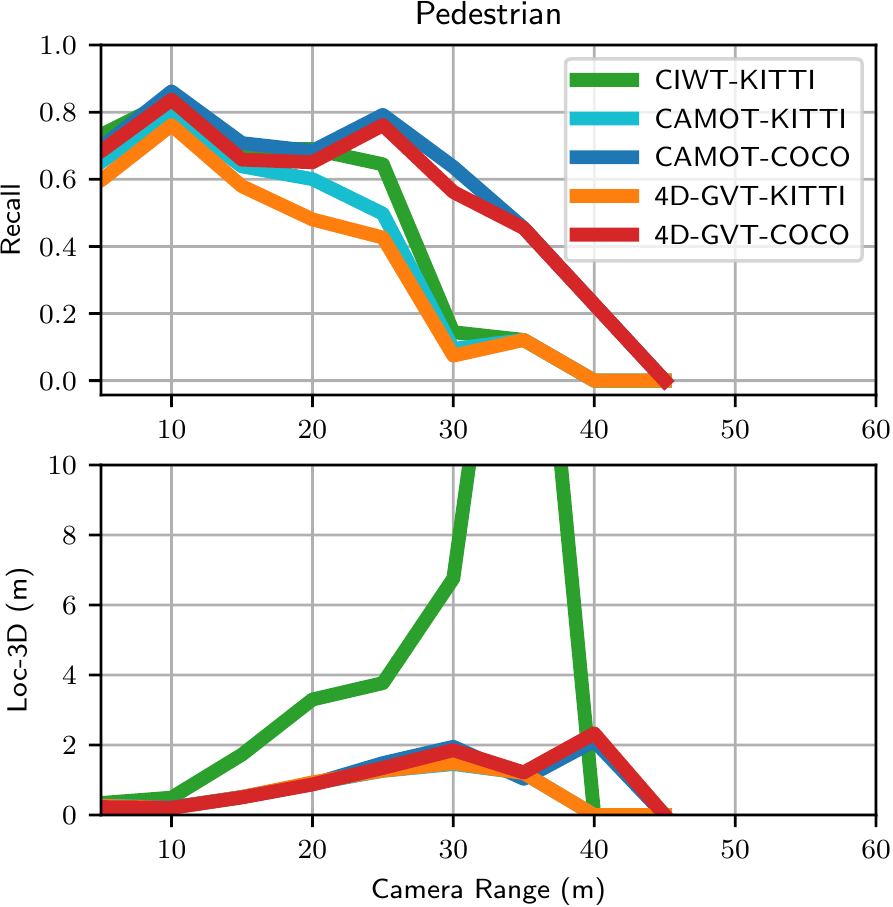}
    \includegraphics[width=0.49\linewidth]{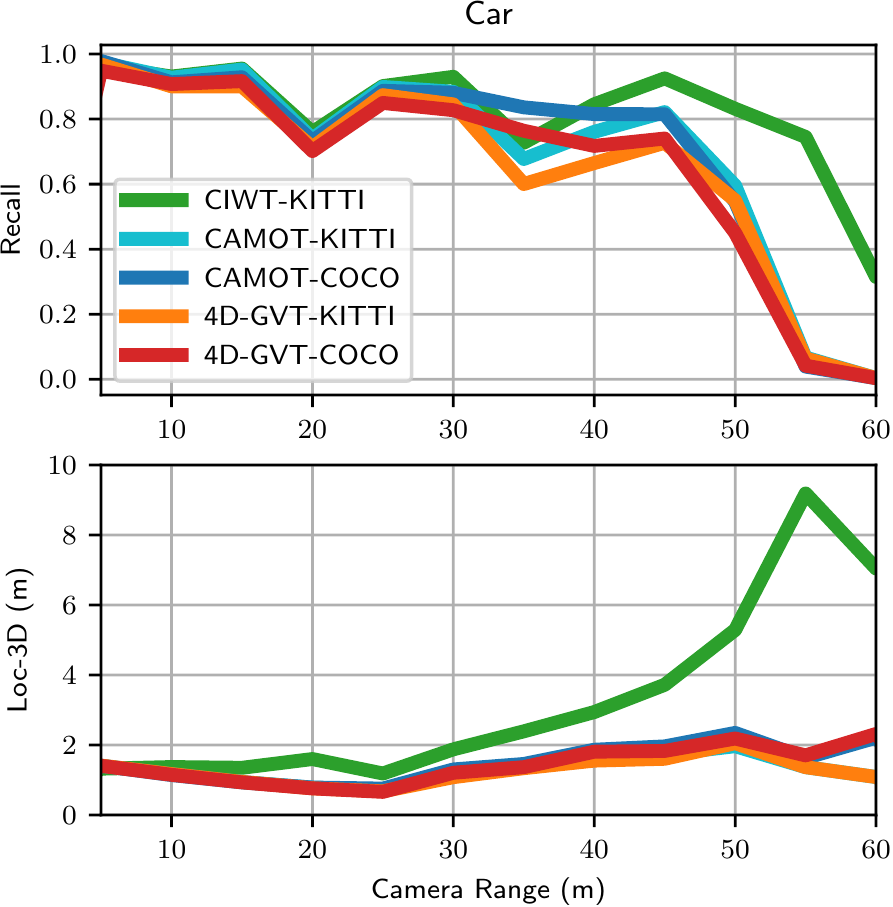}
    \includegraphics[width=0.49\linewidth]{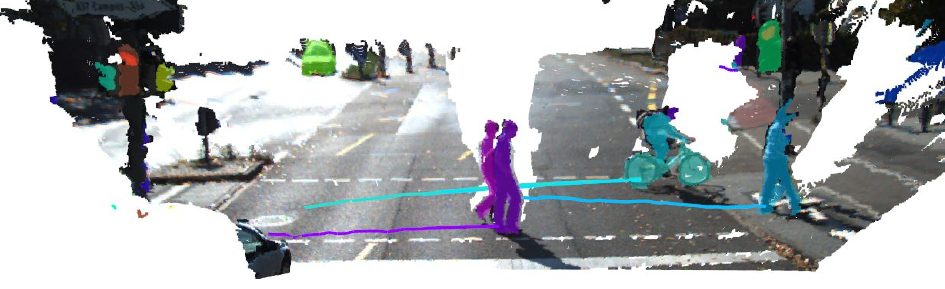}
    \includegraphics[width=0.49\linewidth]{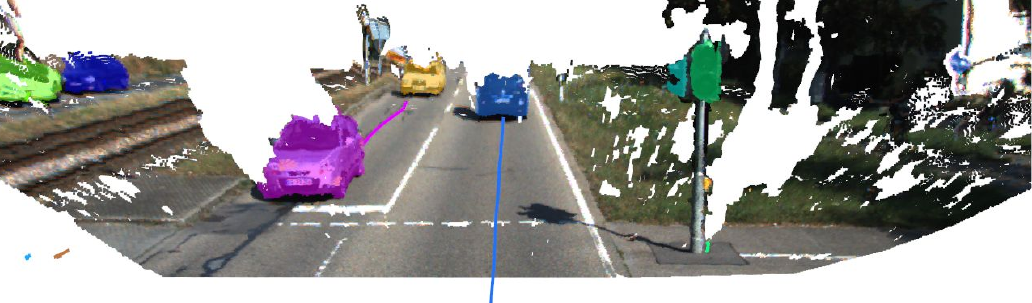}
    \vspace{-17pt}
    \caption{3D localization evaluation on KITTI MOT for \categoryname{pedestrian} (\textit{left}) and \categoryname{car} (\textit{right}). Qualitative results at the bottom.}%
\label{fig:vom-3d-loc}
\end{figure}

Table~\ref{tab:clearmot-ped} outlines the tracking results for the \car and \pedestrian categories. We focus on the MOTA metric and the number of ID switches (IDS). 
CIWT achieves the highest MOTA score ($0.65$) on the \car category when fine-tuned on KITTI, and our 4D-GVT takes the second place ($0.61$). This is due to the higher CIWT recall and that our method can only propose tubes in the near camera range (up to $40-50m$), whereas CIWT can also track in farther ranges in the absence of stereo, as shown in Fig.~\ref{fig:vom-3d-loc}.
For the \pedestrian category, our method and CIWT both achieve $0.33$ MOTA, as KITTI pedestrians mainly appear in the near-range. 
The proposed \fdgvt consistently outperforms CAMOT in terms of MOTA.
As expected, the fine-tuned variants perform better in terms of MOTA score, however, the generic (COCO) model yields better recall in farther camera ranges.
These experiments demonstrate that our video-object proposal generator performs better than a category-agnostic tracker and can compete with top-performing tracking-by-detection methods for \known object categories.
Importantly, our \fdgvt drastically reduces ID switches, which is very important for video object proposals.
It is desirable to extract long video tubes from videos rather than only short fragments.
Comparing the fine-tuned model, \fdgvt commits $18$ and $6$ ID switches for \categoryname{car} and \categoryname{pedestrian}, respectively, compared to $42$ and $8$ (CIWT); $72$ and $40$ (CAMOT).%

Additionally, we evaluate 3D localization precision on KITTI, measured as Euclidean distance (in meters) between estimated 3D positions and ground-truth trajectories. Fig.~\ref{fig:vom-3d-loc} shows recall and 3D localization error as a function of distance from the camera for \categoryname{pedestrians} and \categoryname{cars}.
Both \fdgvt and CAMOT achieve a small localization error -- up to $2m$ in farther camera ranges -- which is not surprising, since they are both based on stereo.
CIWT localizes objects using a stereo-based bottom-up segmentation method and switches to monocular localization when it cannot associate a 3D segment to a detection. 
This way, it can track larger objects (\eg cars) farther than $50$m at the price of significantly larger localization errors. %
This confirms that stereo-based localization is essential for high 3D localization precision.

\PAR{Runtime Analysis.}~With a small number of frame-level proposals (\eg, only cars and pedestrians, 10-20 inputs/frame), \fdgvt requires less than $1$s/frame.
\fdgvt can also operate with a large number of frame-level proposals to provide tubes representing \categoryname{unknown} objects and trade-off between recall/runtime by a varying number of input proposals -- $500$ proposals/frame result in a runtime of $\sim\!\!35$ s/frame. 
This operation mode is intended for offline processing, \eg, collecting tubes of \categoryname{unknown} objects.
Thus, the proposed method is efficient for extracting tubes for  \categoryname{known} objects only, while it is still suitable for mining arbitrary objects from videos offline. 
We performed large-scale video object mining using $14$h of stereo recordings of the Oxford RobotCar dataset. %
In total, we obtained $977,622$ video tubes, of which $833,293$ ($85\%$) are of an \unknown class; these represent potentially interesting novel object categories. %

\newcommand{\mysize}{1.0}
\begin{figure}[t]
    \centering
    \setlength{\fboxsep}{0.3pt}%
    \begin{subfigure}[b]{0.32\linewidth}
        \fbox{\includegraphics[width=\mysize\linewidth]{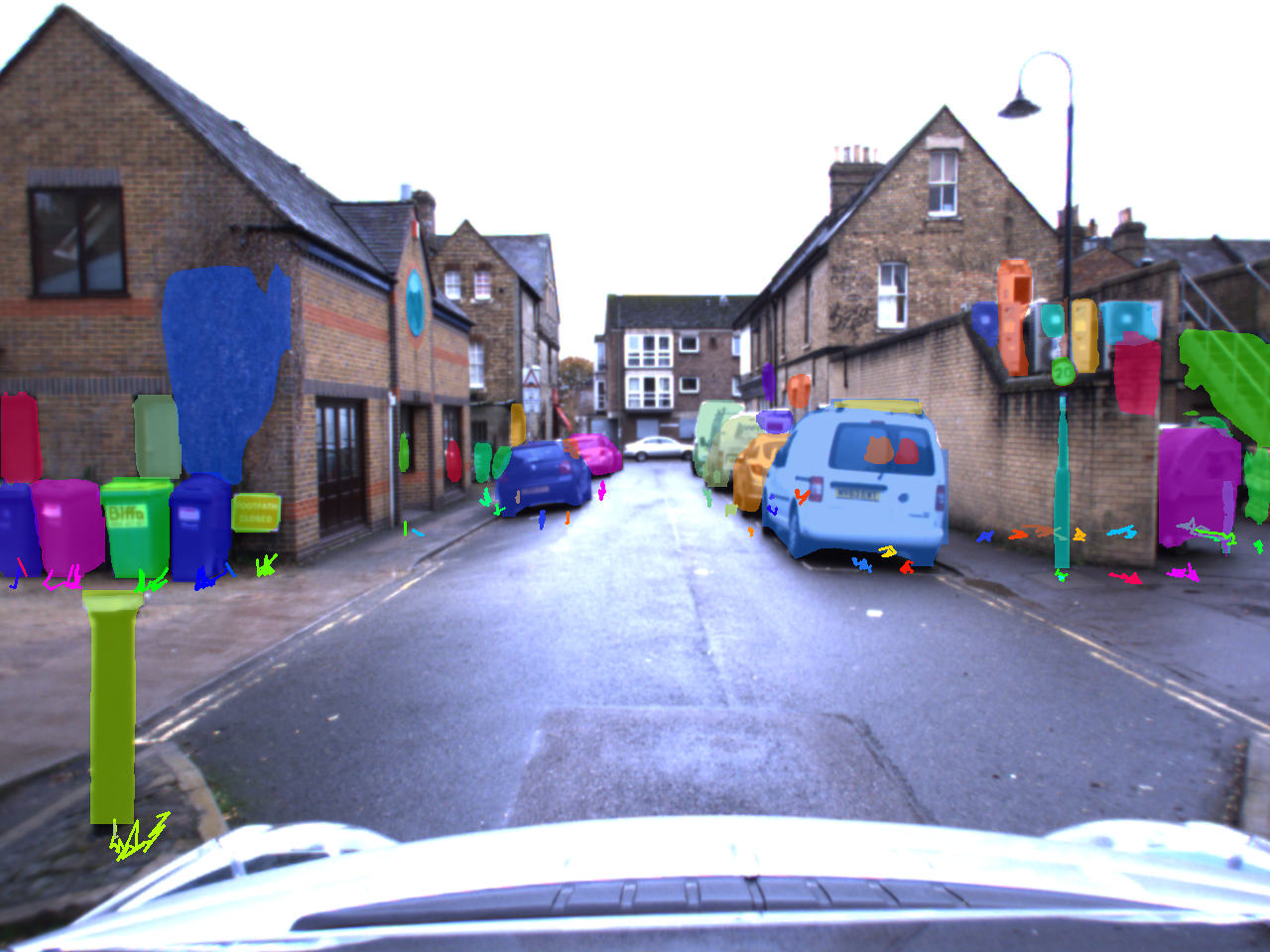}} \\
        \fbox{\includegraphics[width=\mysize\linewidth]{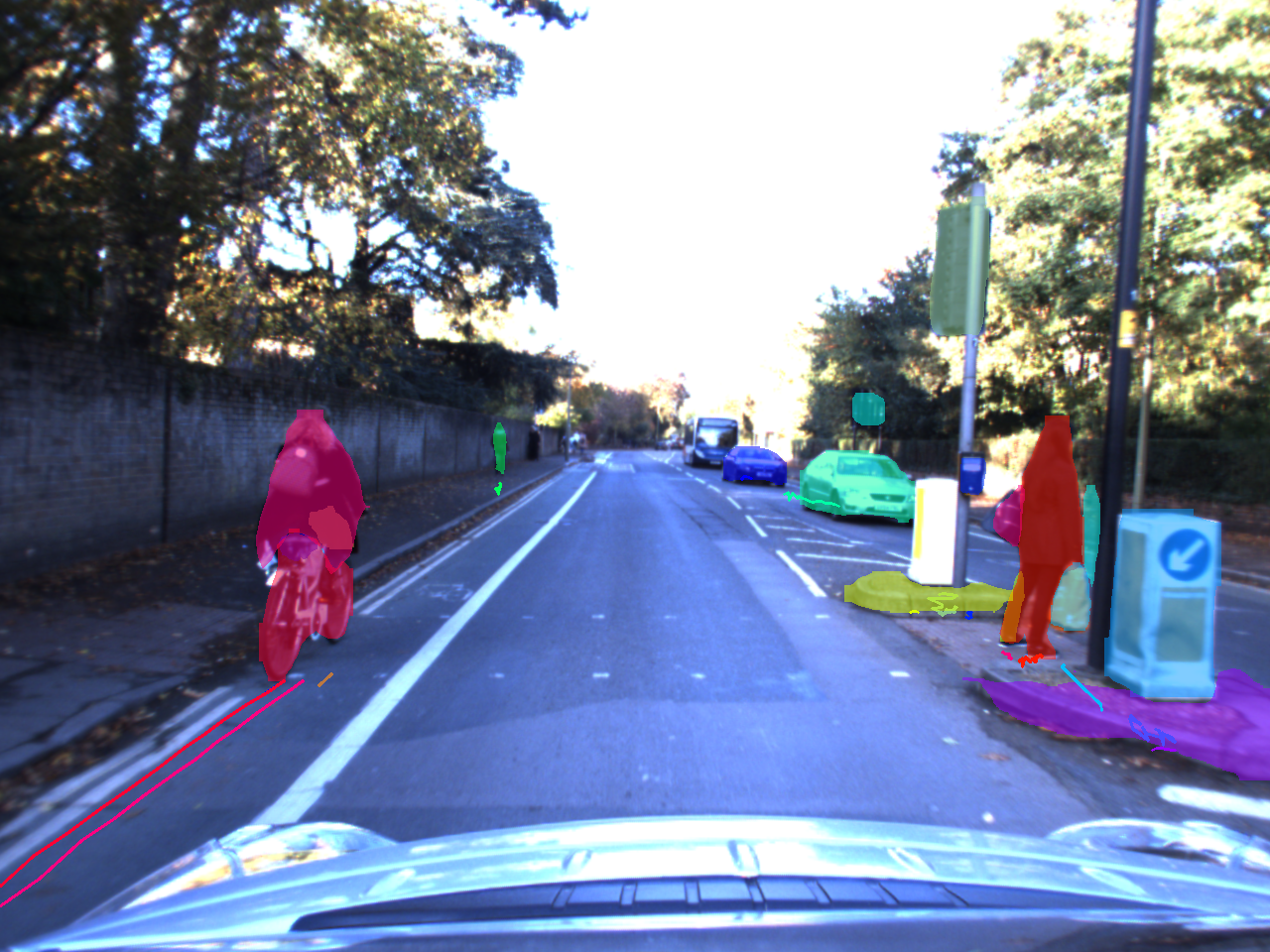}}
        \vspace{-16pt}
        \caption{4D-GVT}
    \end{subfigure}
    \begin{subfigure}[b]{0.32\linewidth}
        \fbox{\includegraphics[width=\mysize\linewidth]{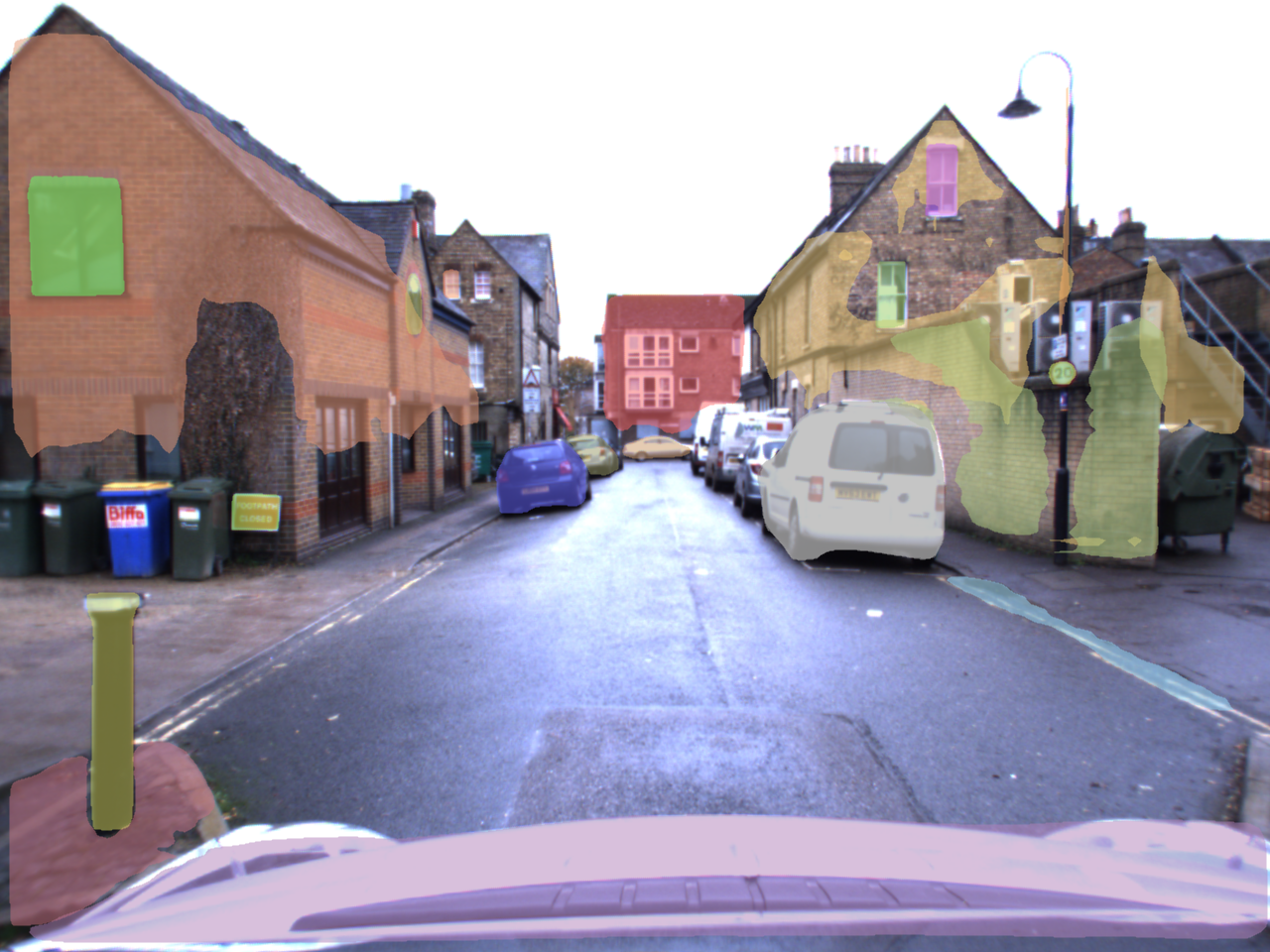}} \\
        \fbox{\includegraphics[width=\mysize\linewidth]{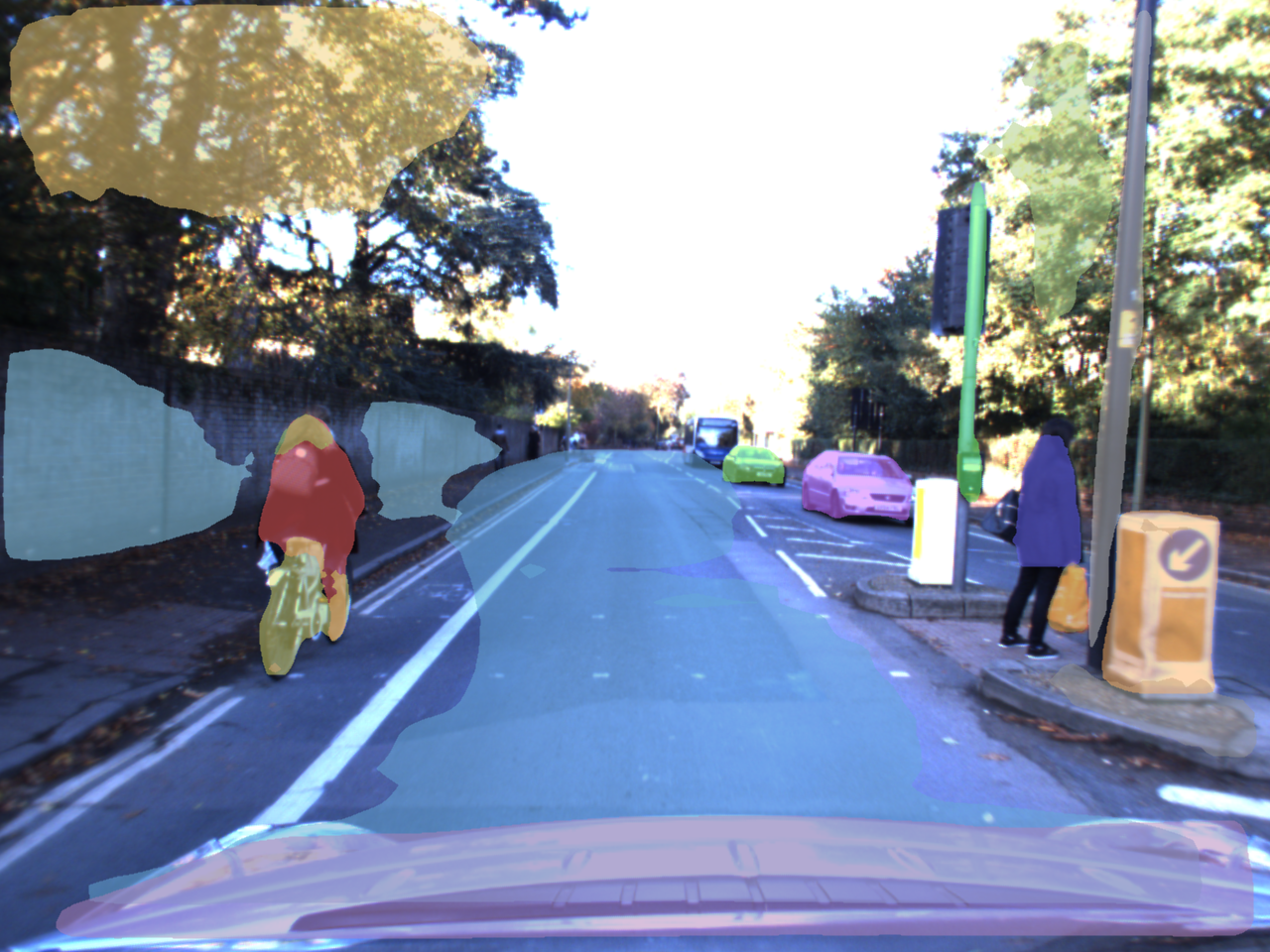}}
        \vspace{-16pt}
        \caption{\maskxrcnn}
    \end{subfigure}
    \begin{subfigure}[b]{0.32\linewidth}
        \fbox{\includegraphics[width=\mysize\linewidth]{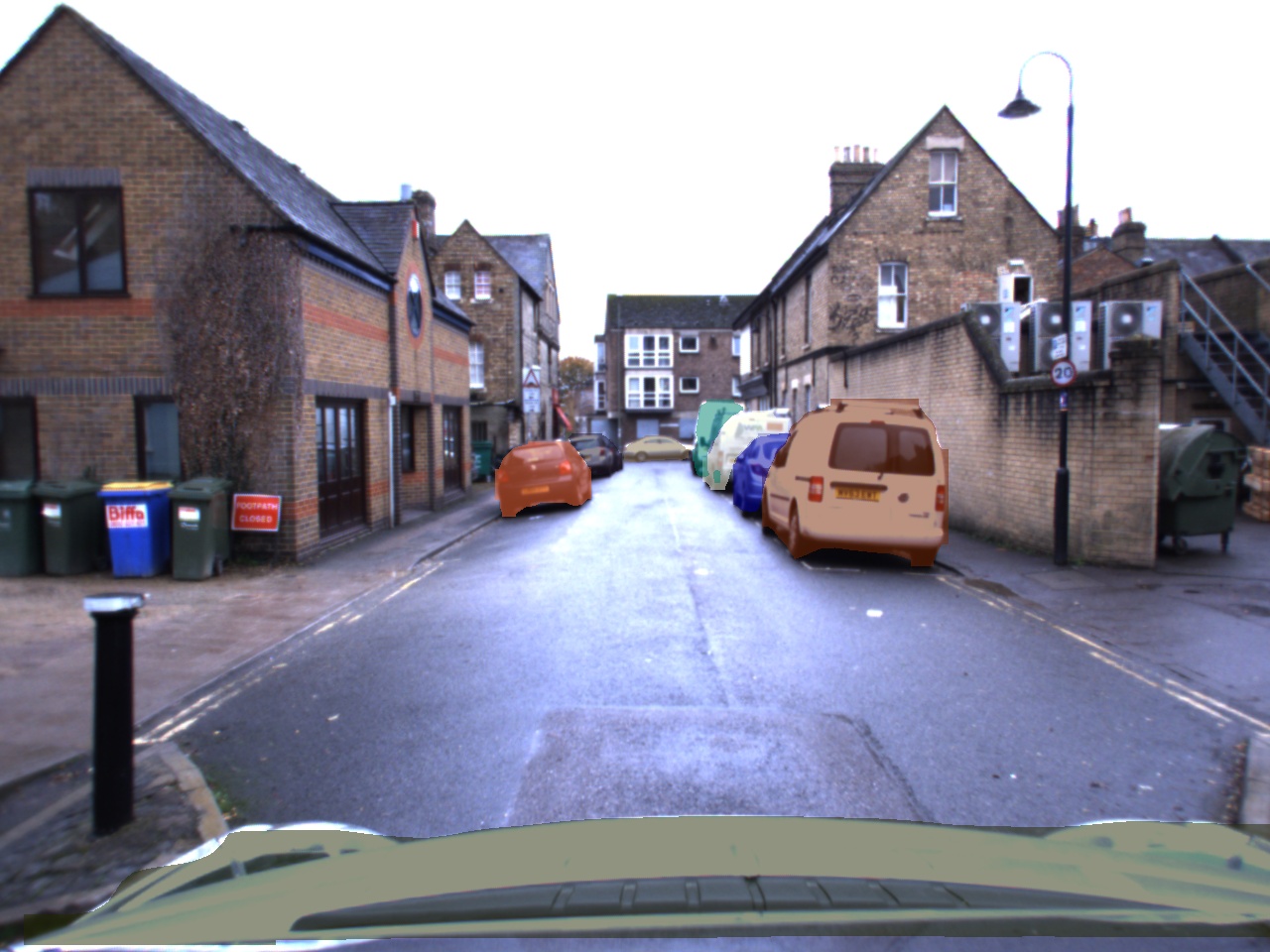}} \\
        \fbox{\includegraphics[width=\mysize\linewidth]{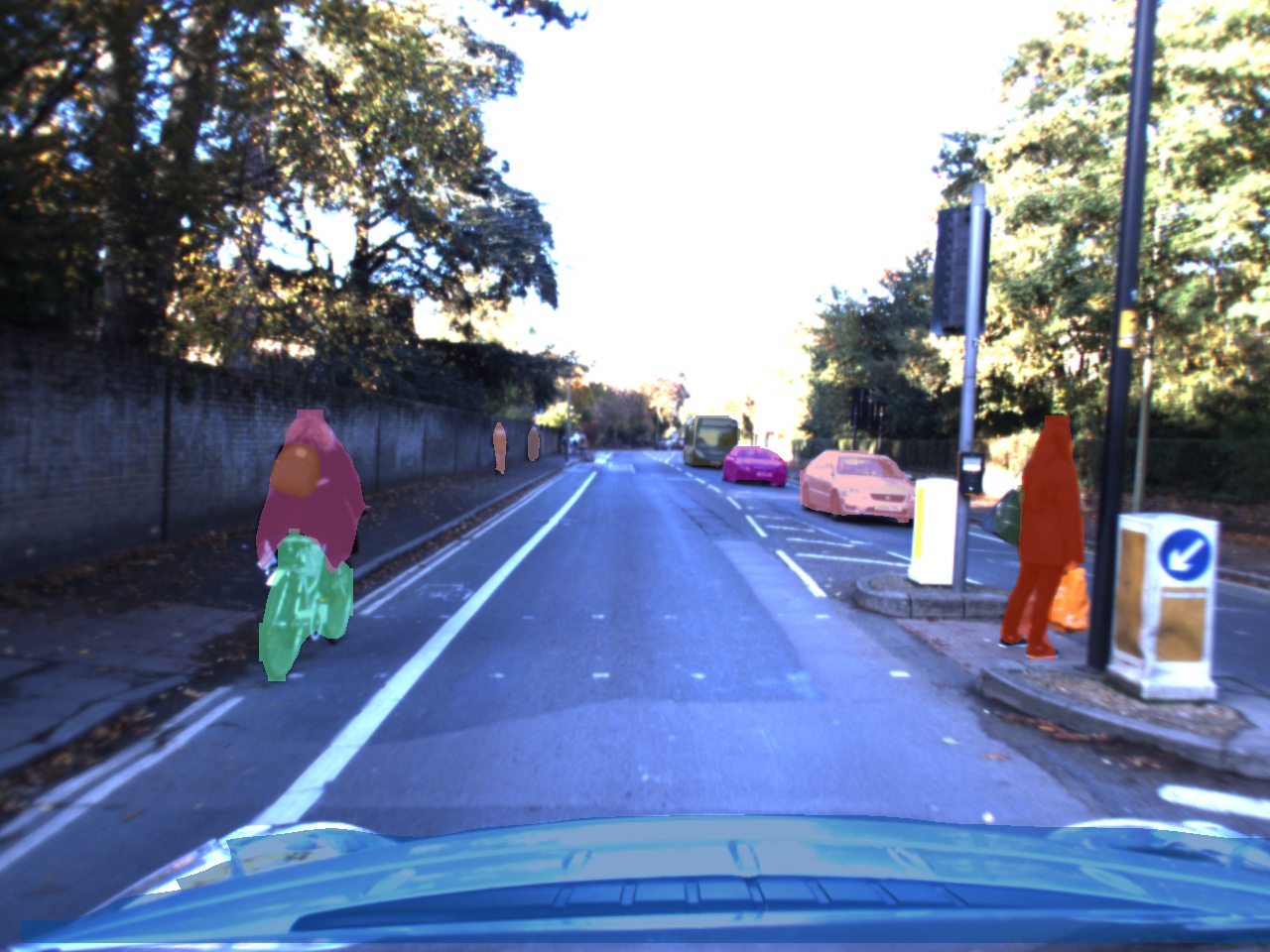}}
        \vspace{-16pt}
        \caption{Mask R-CNN}
    \end{subfigure}
    \vspace{-5pt}
    \caption{Qualitative comparison of the proposed \fdgvt to Mask R-CNN~\cite{He17ICCV} and \maskxrcnn\cite{Hu18CVPR}.}
    \label{fig:qualitative}
\end{figure}

\section{Conclusion}

In this work we have proposed \fdgvt for generic video-object tube generation. 
We have demonstrated that by combining modern instance segmentation methods with a well-funded tracking framework and using parallax as a consistency filter, \fdgvt can match the recall of the recent \maskxrcnn method (trained on a significantly larger dataset) while generating high-quality video object tubes as output that precisely localize objects in 3D.

\footnotesize \PAR{Acknowledgements:} We would like to thank Bin Huang and Michael Krause for annotation work. This project was funded, in parts, by ERC Consolidator Grant DeeVise (ERC-2017-COG-773161). The experiments were performed with computing resources granted by RWTH Aachen University under project rwth0275.

\clearpage
{\small
\bibliographystyle{ieee}
\bibliography{abbrev_short,osep_icra20}

\begin{thebibliography}{10}\itemsep=-1pt

\bibitem{Alahi16CVPR}
A.~Alahi, K.~Goel, V.~Ramanathan, A.~Robicquet, L.~Fei-Fei, and S.~Savarese.
\newblock Social lstm: Human trajectory prediction in crowded spaces.
\newblock In {\em CVPR}, June 2016.

\bibitem{Bansal18ECCV}
A.~Bansal, K.~Sikka, G.~Sharma, R.~Chellappa, and A.~Divakaran.
\newblock Zero-shot object detection.
\newblock In {\em ECCV}, 2018.

\bibitem{Bernardin08JIVP}
K.~Bernardin and R.~Stiefelhagen.
\newblock Evaluating multiple object tracking performance: The clear mot
  metrics.
\newblock {\em JVIP}, 2008:1:1--1:10, 2008.

\bibitem{Brendel12ICCV}
W.~Brendel and S.~Todorovic.
\newblock Video object segmentation by tracking regions.
\newblock In {\em ICCV}, 2009.

\bibitem{Choi15ICCV}
W.~Choi.
\newblock Near-online multi-target tracking with aggregated local flow
  descriptor.
\newblock In {\em ICCV}, 2015.

\bibitem{Dewan15ICRA}
A.~Dewan, T.~Caselitz, G.~D. Tipaldi, and W.~Burgard.
\newblock Motion-based detection and tracking in 3d lidar scans.
\newblock In {\em ICRA}, 2015.

\bibitem{Fragkiadaki15CVPR}
K.~Fragkiadaki, P.~A. Arbel{\'a}ez, P.~Felsen, and J.~Malik.
\newblock Learning to segment moving objects in videos.
\newblock {\em CVPR}, 2015.

\bibitem{Geiger12CVPR}
A.~Geiger, P.~Lenz, and R.~Urtasun.
\newblock Are we ready for autonomous driving? the {KITTI} vision benchmark
  suite.
\newblock In {\em CVPR}, 2012.

\bibitem{Geiger10ACCV}
A.~Geiger, M.~Roser, and R.~Urtasun.
\newblock Efficient large-scale stereo matching.
\newblock In {\em ACCV}, 2010.

\bibitem{Geiger11IV}
A.~Geiger, J.~Ziegler, and C.~Stiller.
\newblock Stereoscan: Dense {3D} reconstruction in real-time.
\newblock In {\em Intel. Vehicles Symp.}, 2011.

\bibitem{Gupta18CVPR}
A.~Gupta, J.~Johnson, L.~Fei-Fei, S.~Savarese, and A.~Alahi.
\newblock Social gan: Socially acceptable trajectories with generative
  adversarial networks.
\newblock In {\em CVPR}, 2018.

\bibitem{He17ICCV}
K.~He, G.~Gkioxari, P.~Doll\'{a}r, and R.~Girshick.
\newblock {Mask R-CNN}.
\newblock In {\em ICCV}, 2017.

\bibitem{Held16RSS}
D.~Held, D.~Guillory, B.~Rebsamen, S.~Thrun, and S.~Savarese.
\newblock A probabilistic framework for real-time 3d segmentation using
  spatial, temporal, and semantic cues.
\newblock In {\em RSS}, 2016.

\bibitem{Held14RSS}
D.~Held, J.~Levinson, S.~Thrun, and S.~Savarese.
\newblock Combining 3d shape, color, and motion for robust anytime tracking.
\newblock In {\em RSS}, 2014.

\bibitem{Horbert15ICRA}
E.~Horbert, G.~M. Garcia, S.~Frintrop, and B.~Leibe.
\newblock Sequence-level object candidates based on saliency for generic object
  recognition on mobile systems.
\newblock In {\em ICRA}, 2015.

\bibitem{Hou17ICCV}
R.~Hou, C.~Chen, and M.~Shah.
\newblock Tube convolutional neural network (t-cnn) for action detection in
  videos.
\newblock In {\em ICCV}, 2017.

\bibitem{Hu18CVPR}
R.~Hu, P.~Doll\'{a}r, K.~He, T.~Darrell, and R.~Girshick.
\newblock {Learning to Segment Every Thing}.
\newblock In {\em CVPR}, 2018.

\bibitem{Ting17NIPS}
Y.~Hu, J.~Huang, and A.~Schwing.
\newblock Maskrnn: Instance level video object segmentation.
\newblock In {\em NIPS}, 2017.

\bibitem{Jin18ECCV}
S.~Jin, A.~RoyChowdhury, H.~Jiang, A.~Singh, A.~Prasad, D.~Chakraborty, and
  E.~Learned-Miller.
\newblock Unsupervised hard example mining from videos for improved object
  detection.
\newblock In {\em ECCV}, 2018.

\bibitem{Krishna16Arxiv}
R.~Krishna, Y.~Zhu, O.~Groth, J.~Johnson, K.~Hata, J.~Kravitz, S.~Chen,
  Y.~Kalantidis, L.-J. Li, D.~A. Shamma, M.~Bernstein, and L.~Fei-Fei.
\newblock Visual genome: Connecting language and vision using crowdsourced
  dense image annotations.
\newblock {\em arXiv preprint arXiv:1602.07332}, 2016.

\bibitem{Kwak15ICCV}
S.~Kwak, M.~Cho, I.~Laptev, J.~Ponce, and C.~Schmid.
\newblock Unsupervised object discovery and tracking in video collections.
\newblock In {\em ICCV}, 2015.

\bibitem{Lee17CVPR}
N.~Lee, W.~Choi, P.~Vernaza, C.~B. Choy, P.~H.~S. Torr, and M.~K. Chandraker.
\newblock Desire: Distant future prediction in dynamic scenes with interacting
  agents.
\newblock {\em CVPR}, 2017.

\bibitem{Lee10CVPR}
Y.~J. Lee and K.~Grauman.
\newblock Object-graphs for context-aware visual category discovery.
\newblock In {\em CVPR}, 2010.

\bibitem{Lee11CVPR}
Y.~J. Lee and K.~Grauman.
\newblock Learning the easy things first: Self-paced visual category discovery.
\newblock In {\em CVPR}, 2011.

\bibitem{Leibe08TPAMI}
B.~Leibe, K.~Schindler, N.~Cornelis, and L.~V. Gool.
\newblock Coupled object detection and tracking from static cameras and moving
  vehicles.
\newblock {\em PAMI}, 30(10):1683--1698, 2008.

\bibitem{Lenz11IV}
P.~Lenz, J.~Ziegler, A.~Geiger, and M.~Roser.
\newblock Sparse scene flow segmentation for moving object detection in urban
  environments.
\newblock In {\em Intel. Vehicles Symp.}, 2011.

\bibitem{Li18CVPR}
S.~Li, B.~Seybold, A.~Vorobyov, A.~Fathi, Q.~Huang, and C.-C.~J. Kuo.
\newblock Instance embedding transfer to unsupervised video object
  segmentation.
\newblock In {\em CVPR}, 2018.

\bibitem{Lin14ECCV}
T.-Y. Lin, M.~Maire, S.~Belongie, J.~Hays, P.~Perona, D.~Ramanan,
  P.~Doll{\'a}r, and C.~L. Zitnick.
\newblock Microsoft {COCO}: Common objects in context.
\newblock In {\em ECCV}, 2014.

\bibitem{Maddern17IJRR}
W.~Maddern, G.~Pascoe, C.~Linegar, and P.~Newman.
\newblock 1 year, 1000km: The {Oxford RobotCar} dataset.
\newblock {\em IJRR}, 36(1):3--15, 2017.

\bibitem{Misra15CVPR}
I.~Misra, A.~Shrivastava, and M.~Hebert.
\newblock Watch and learn: Semi-supervised learning of object detectors from
  videos.
\newblock In {\em CVPR}, 2015.

\bibitem{Misra16ECCV}
I.~Misra, C.~L. Zitnick, and M.~Hebert.
\newblock {Shuffle and Learn}: {Unsupervised} learning using temporal order
  verification.
\newblock In {\em ECCV}, 2016.

\bibitem{Mitzel12ECCV}
D.~Mitzel and B.~Leibe.
\newblock Taking mobile multi-object tracking to the next level: People,
  unknown objects, and carried items.
\newblock In {\em ECCV}, 2012.

\bibitem{Osep16ICRA}
A.~O\v{s}ep, A.~Hermans, F.~Engelmann, D.~Klostermann, M.~Mathias, and
  B.~Leibe.
\newblock Multi-scale object candidates for generic object tracking in street
  scenes.
\newblock In {\em ICRA}, 2016.

\bibitem{Osep17ICRA}
A.~O\v{s}ep, W.~Mehner, M.~Mathias, and B.~Leibe.
\newblock Combined image- and world-space tracking in traffic scenes.
\newblock In {\em ICRA}, 2017.

\bibitem{Osep18ICRA}
A.~O\v{s}ep, W.~Mehner, P.~Voigtlaender, and B.~Leibe.
\newblock Track, then decide: Category-agnostic vision-based multi-object
  tracking.
\newblock {\em ICRA}, 2018.

\bibitem{Pham18ECCV}
T.~Pham, V.~B.~G. Kumar, T.-T. Do, G.~Carneiro, and I.~Reid.
\newblock Bayesian semantic instance segmentation in open set world.
\newblock In {\em ECCV}, September 2018.

\bibitem{Pinheiro16ECCV}
P.~Pinheiro, T.~Lin, R.~Collobert, and P.~Doll{\'{a}}r.
\newblock Learning to refine object segments.
\newblock In {\em ECCV}, 2016.

\bibitem{Prest12CVPR}
A.~Prest, C.~Leistner, J.~Civera, C.~Schmid, and V.~Ferrari.
\newblock Learning object class detectors from weakly annotated video.
\newblock In {\em CVPR}, 2012.

\bibitem{Rahman18ACCV}
S.~Rahman, S.~H. Khan, and F.~Porikli.
\newblock Zero-shot object detection: Learning to simultaneously recognize and
  localize novel concepts.
\newblock {\em ACCV}, 2018.

\bibitem{Reid79TAC}
D.~B. Reid.
\newblock An algorithm for tracking multiple targets.
\newblock {\em IEEE Trans. Automatic Control}, 24(6):843--854, 1979.

\bibitem{Rubinstein13CVPR}
M.~Rubinstein, A.~Joulin, J.~Kopf, and C.~Liu.
\newblock Unsupervised joint object discovery and segmentation in internet
  images.
\newblock In {\em CVPR}, 2013.

\bibitem{Russell06CVPR}
B.~C. Russell, W.~T. Freeman, A.~A. Efros, J.~Sivic, and A.~Zisserman.
\newblock Using multiple segmentations to discover objects and their extent in
  image collections.
\newblock In {\em CVPR}, 2006.

\bibitem{Schindler06ECCV}
K.~Schindler, J.~U., and H.~Wang.
\newblock Perspective {N-view} multibody structure-and-motion through model
  selection.
\newblock In {\em ECCV}, 2006.

\bibitem{Seguin16CVPR}
G.~Seguin, P.~Bojanowski, R.~Lajugie, and I.~Laptev.
\newblock Instance-level video segmentation from object tracks.
\newblock In {\em CVPR}, 2016.

\bibitem{Sivic08CVPR}
J.~Sivic, B.~C. Russell, A.~Zisserman, W.~T. Freeman, and A.~A. Efros.
\newblock Unsupervised discovery of visual object class hierarchies.
\newblock In {\em CVPR}, 2008.

\bibitem{Stutz18CVPR}
D.~Stutz and A.~Geiger.
\newblock Learning 3d shape completion from laser scan data with weak
  supervision.
\newblock In {\em CVPR}, 2018.

\bibitem{Tang12NIPS}
K.~Tang, V.~Ramanathan, L.~Fei-fei, and D.~Koller.
\newblock Shifting weights: Adapting object detectors from image to video.
\newblock In {\em NIPS}, 2012.

\bibitem{Teichman11ICRA}
A.~Teichman, J.~Levinson, and S.~Thrun.
\newblock Towards {3D} object recognition via classification of arbitrary
  object tracks.
\newblock In {\em ICRA}, 2011.

\bibitem{Teichman12IJRR}
A.~Teichman and S.~Thrun.
\newblock Tracking-based semi-supervised learning.
\newblock {\em IJRR}, 31(7):804--818, 2012.

\bibitem{Tokmakov17ICCV}
P.~Tokmakov, K.~Alahari, and C.~Schmid.
\newblock Learning video object segmentation with visual memory.
\newblock In {\em ICCV}, 2017.

\bibitem{Tokmakov18IJCV}
P.~Tokmakov, C.~Schmid, and K.~Alahari.
\newblock Learning to segment moving objects.
\newblock {\em IJCV}, 127:282--301, 2018.

\bibitem{Tuytelaars10IJCV}
T.~Tuytelaars, C.~H. Lampert, M.~B. Blaschko, and W.~Buntine.
\newblock Unsupervised object discovery: A comparison.
\newblock {\em IJCV}, 88:284--302, 2010.

\bibitem{Vogel13ICCV}
C.~Vogel, K.~Schindler, and S.~Roth.
\newblock Piecewise rigid scene flow.
\newblock In {\em ICCV}, 2013.

\bibitem{Voigtlaender17BMVC}
P.~Voigtlaender and B.~Leibe.
\newblock Online adaptation of convolutional neural networks for video object
  segmentation.
\newblock In {\em BMVC}, 2017.

\bibitem{Vondrick18ECCV}
C.~Vondrick, A.~Shrivastava, A.~Fathi, S.~Guadarrama, and K.~Murphy.
\newblock Tracking emerges by colorizing videos.
\newblock In {\em ECCV}, September 2018.

\bibitem{Wang14TPAMI}
L.~Wang, G.~Hua, R.~Sukthankar, J.~Xue, Z.~Niu, and N.~Zheng.
\newblock Video object discovery and co-segmentation with extremely weak
  supervision.
\newblock {\em PAMI}, 39:2074--2088, 2014.

\bibitem{Wang13ICCV}
X.~Wang, M.~Yang, S.~Zhu, and Y.~Lin.
\newblock Regionlets for generic object detection.
\newblock In {\em ICCV}, 2013.

\bibitem{Xian18TPAMI}
Y.~Xian, C.~H. Lampert, B.~Schiele, and Z.~Akata.
\newblock Zero-shot learning - a comprehensive evaluation of the good, the bad
  and the ugly.
\newblock {\em PAMI}, 2018.

\bibitem{Yoon16CVPR}
J.~H. Yoon, C.-R. Lee, M.-H. Yang, and K.-J. Yoon.
\newblock Online multi-object tracking via structural constraint event
  aggregation.
\newblock In {\em CVPR}, 2016.

\bibitem{Yuan183DV}
W.~Yuan, T.~Khot, D.~Held, C.~Mertz, and M.~Hebert.
\newblock Pcn: Point completion network.
\newblock In {\em 3DV}, 2018.

\bibitem{Zhang08CVPR}
L.~Zhang, L.~Yuan, and R.~Nevatia.
\newblock Global data association for multi-object tracking using network
  flows.
\newblock In {\em CVPR}, 2008.

\bibitem{Zhu16ACCV}
G.~Zhu, F.~Porikli, and H.~Li.
\newblock Model-free multiple object tracking with shared proposals.
\newblock In {\em ACCV}, 2016.

\bibitem{Zhu12CVPR}
J.-Y. Zhu, J.~Wu, Y.~Wei, E.~Chang, and Z.~Tu.
\newblock Unsupervised object class discovery via saliency-guided multiple
  class learning.
\newblock In {\em CVPR}, 2012.

\bibitem{Zhu19TCSV}
P.~{Zhu}, H.~{Wang}, and V.~{Saligrama}.
\newblock Zero shot detection.
\newblock {\em IEEE Trans. on Circ. and Systems for Video Tech.}, pages 1--1,
  2019.

\end{thebibliography}
}

\end{document}